\documentclass{article}
\usepackage[accepted]{icml2024}

\usepackage{microtype}
\usepackage{graphicx}
\usepackage{subfigure}
\usepackage{booktabs} %

\usepackage[colorlinks=true,linkcolor=blue,urlcolor=blue,citecolor=teal]{hyperref}

\usepackage{pythonhighlight}
\usepackage{subcaption}
\usepackage{graphicx}
\usepackage{amssymb}
\usepackage{wrapfig}
\usepackage{multicol}
\usepackage{xcolor}
\usepackage{mathtools}
\usepackage[ruled,noend,algo2e]{algorithm2e}
\SetKwInput{KwInput}{Input}
\usepackage{multirow}
\usepackage{booktabs}
\usepackage{subfigure}
\usepackage{subcaption}
\usepackage{tcolorbox}
\usepackage{enumitem}
\usepackage{lipsum}
\usepackage{adjustbox}
\usepackage{calc}
\usepackage{caption}
\usepackage{subcaption}

\newlength{\subcolumnwidth}

\newcommand{\nextsubcolumn}[1][]{%
  \cr\noalign{\hfill}
  \if\relax\detokenize{#1}\relax\else\hsize=#1\setlength{\subcolumnwidth}{\hsize}\fi
}

\usepackage{amsmath}
\usepackage{amssymb}
\usepackage{mathtools}
\usepackage{amsthm}

\usepackage[capitalize,noabbrev]{cleveref}

\theoremstyle{plain}

\theoremstyle{definition}

\theoremstyle{remark}

\usepackage[textsize=tiny]{todonotes}

\icmltitlerunning{LTE: Parallel Low-Rank Updates}

\newcommand{\fig}[1]{Figure~\ref{#1}}
\newcommand{\sect}[1]{Section~\ref{#1}}
\newcommand{\tbl}[1]{Table~\ref{#1}}

\newcommand{\app}[1]{Appendix~\ref{#1}}
\newcommand{\alg}[1]{Algorithm~\ref{#1}}

\newcommand{\ignorethis}[1]{}

\usepackage{xcolor}
\usepackage{amsthm}
\usepackage{framed}

\newcommand{\xpar}[1]{\noindent\textbf{#1}\ \ }

\newcommand{\comm}[1]{}

\newcommand{\cL}{\mathcal{L}}

\newcommand\blfootnote[1]{%
  \begingroup
  \renewcommand\thefootnote{}\footnote{#1}%
  \addtocounter{footnote}{-1}%
  \endgroup
}

\newcommand{\bx}{\mathbf{x}}
\newcommand{\by}{\mathbf{y}}
\newcommand{\bz}{\mathbf{z}}

\newcommand{\bW}{\mathbf{W}}
\newcommand{\bA}{\mathbf{A}}
\newcommand{\bB}{\mathbf{B}}
\newcommand{\bX}{\mathbf{X}}
\newcommand{\bY}{\mathbf{Y}}
\newcommand{\bV}{\mathbf{V}}

\newcommand{\bb}{\mathbf{b}}
\newcommand{\ba}{\mathbf{a}}

\DeclareMathOperator*{\argmin}{arg\,min}

\makeatletter
\def\blfootnote{\gdef\@thefnmark{}\@footnotetext}
\makeatother

\begin{document}

\twocolumn[
\icmltitle{Training Neural Networks from Scratch with Parallel Low-Rank Adapters}

\begin{icmlauthorlist}
\icmlauthor{Minyoung Huh}{mit_csail}
\icmlauthor{Brian Cheung}{mit_csail,mit_bcs}
\icmlauthor{Jeremy Bernstein}{mit_csail}
\icmlauthor{Phillip Isola}{mit_csail}
\icmlauthor{Pulkit Agrawal}{mit_csail}

\end{icmlauthorlist}

\icmlaffiliation{mit_csail}{MIT CSAIL}
\icmlaffiliation{mit_bcs}{MIT CBMM}

\icmlcorrespondingauthor{Minyoung Huh}{minhuh@mit.edu}

\icmlkeywords{Machine Learning, ICML}

\vskip 0.3in
]

\begin{abstract}
The scalability of deep learning models is fundamentally limited by computing resources, memory, and communication. Although methods like low-rank adaptation (LoRA) have reduced the cost of model finetuning, its application in model pre-training remains largely unexplored. This paper explores extending LoRA to model pre-training, identifying the inherent constraints and limitations of standard LoRA in this context. We introduce \textit{LoRA-the-Explorer} (LTE), a novel bi-level optimization algorithm designed to enable parallel training of multiple low-rank heads across computing nodes, thereby reducing the need for frequent synchronization. Our approach includes extensive experimentation on vision transformers using various vision datasets, demonstrating that LTE is competitive with standard pre-training. 
\vspace{0.1in}\\
$\triangleright \; \mathsf{project\;page}:$ {\small \url{minyoungg.github.io/LTE}}

\end{abstract}

\section{Introduction}
\vspace{0.1in}

The escalating complexity of state-of-the-art deep learning models presents significant challenges not only in terms of computational demand but also in terms of memory and communication bandwidth. As these demands exceed the capacity of consumer-grade GPUs, training larger models requires innovative solutions. A prominent example of finetuning models is low-rank adaptation~\cite{hu2021lora} that uses a low-rank parameterization of a deep neural network to reduce memory requirements for storing optimizer-state and gradient communication during training.
The memory requirement was further reduced by quantizing model parameters~\citep{dettmers2023qlora}. Such innovations have enabled the finetuning of large models, even on a single consumer-grade GPU. 

However, prior work has been limited to finetuning, and tools to pre-train models from scratch are still absent. Hence, the goal of this paper is to extend adaptation methods to model pre-training. Specifically, we ask the question: \textit{Can neural networks be trained from scratch using low-rank adapters?}

Successfully addressing this question carries substantial implications, especially considering that common computing clusters often have slower cross-node training than single-node training with gradient accumulation due to slow communication speed and bandwidth.
Low-rank adapters effectively compress the communication between these processors while preserving essential structural attributes for effective model training. Our investigation reveals that while vanilla LoRA underperforms in training a model from scratch, the use of parallel low-rank updates can bridge this performance gap. 

\begin{figure}[t!]
\centering
    \includegraphics[width=\linewidth]{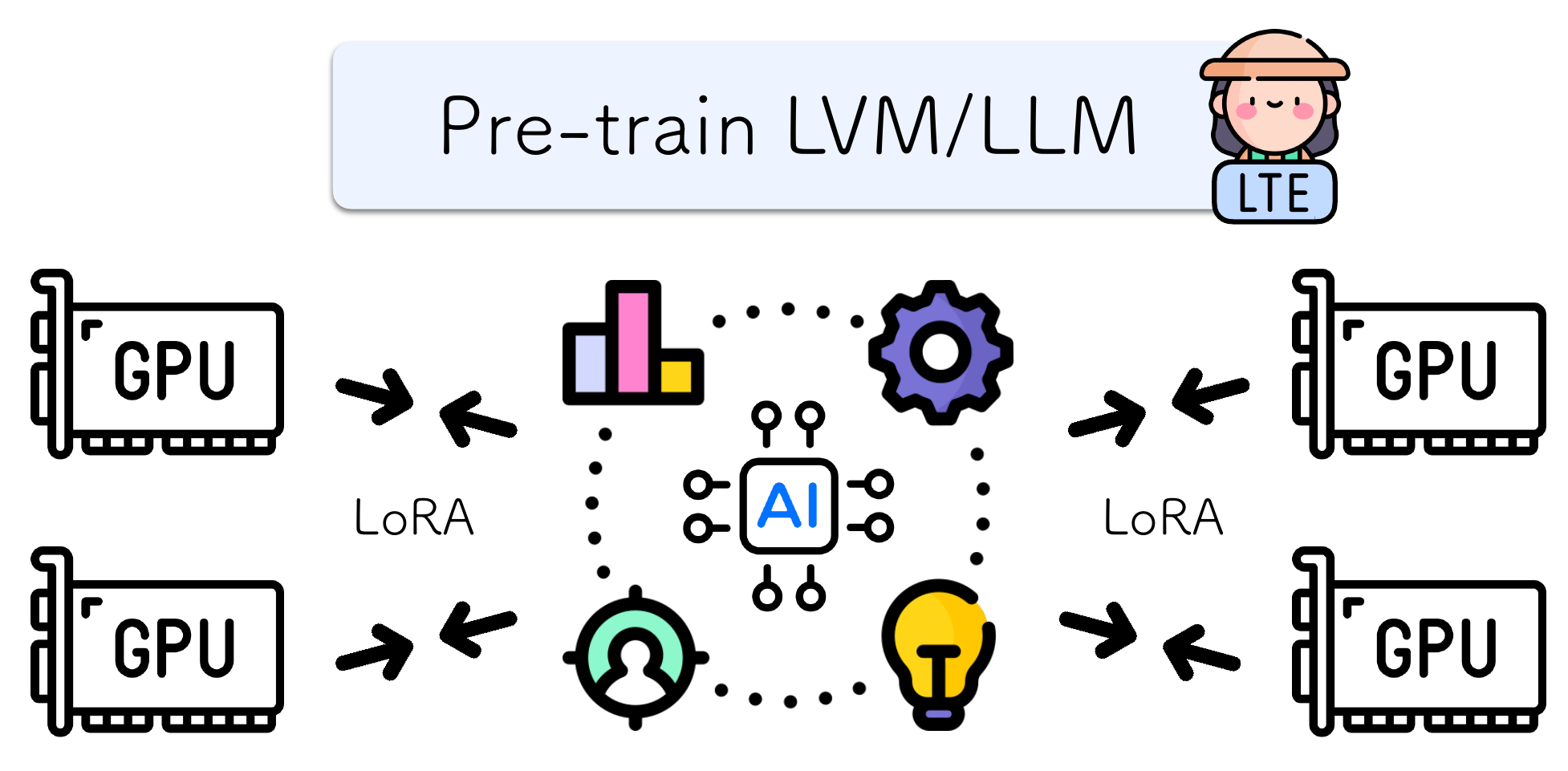}
    \caption{\small\textbf{Lora-The-Explorer:} We propose LoRA-the-explorer, an optimization algorithm that can match the performance of standard training from scratch. Our method optimizes unique LoRA parameters in parallel and merges them back to the main weights. Our algorithm can leverage lower-memory devices and only depends on communicating the LoRA parameters for training, making it an ideal candidate in a bandlimited or memory-constraint training framework.\protect\footnotemark
    }
    \label{fig:full_model_vs_lora}
\end{figure}

\footnotetext{Figure generated using assets from Flaticon~\url{flaticon.com}}

\paragraph{Principal findings and contributions:}
\begin{itemize}[itemsep=0.2pt,topsep=0pt,leftmargin=*]
\item In~\sect{sec:methods}, we establish limitations inherent to LoRA for model pre-training. We show that parallel updates are needed and introduce our algorithmic approach LTE in~\sect{sec:motivation} and~\sect{sec:lte}.
\item In~\sect{sec:experiments}, we combine LTE with federated averaging~\cite{mcmahan2017communication} to demonstrate comparable performance to distributed pre-training, even with infrequent synchronization.
\item We provide empirical analysis and ablation studies in~\sect{sec:merging},~\sect{sec:lora-alignment}, and~\sect{sec:ablation}.
\item In~\sect{sec:imagenet1k}, we conduct a resource utilization comparison with standard distributed data-parallel (DDP) training on $8$ GPUs. Although our method extends the training samples required for convergence by $40\%$, we can fit models that are $3\times$ bigger with roughly half the bandwidth. This, in turn, enables faster training if one has access to many low-memory devices.
\item We provide discussion of related works in~\sect{sec:related}.

\end{itemize}

\section{Preliminaries}
\label{sec:preliminary}

Unless stated otherwise, we denote $x$ as a scalar, $\bx$ a vector, $\bX$ a matrix, $\mathcal{X}$ a distribution or a set, $f(\cdot)$ a function and $F(\cdot)$ a composition of functions, and $\mathcal{L}(\cdot,\cdot)$ a loss-function.

\subsection{Parameter efficient adapters}

Adapters serve as trainable functions that modify existing layers in a neural network. They facilitate parameter-efficient finetuning of large-scale models by minimizing the memory requirements for optimization (see~\sect{sec:related} for various types of adapters used in prior works).

The focus of this work is on the \textit{low-rank adapter}~\citep[LoRA]{hu2021lora}, a subclass of linear adapters. The linearity of LoRA allows for the trained parameters to be integrated back into the existing weights post-training without further tuning or approximation. Hence, the linearity allows models to maintain the original inference cost. LoRA is frequently used for finetuning transformers, often resulting in less than $10\%$ of the total trainable parameters (even as low as $0.5\%$).

\xpar{Low-Rank Adapter (LoRA)} Given input $\bx \in \mathbb{R}^{n}$, and a linear layer $f(\cdot): \mathbb{R}^{n} \rightarrow \mathbb{R}^{m}$ parameterized by the weight $\bW \in \mathbb{R}^{m \times n}$, LoRA re-parameterizes the function as:
\begin{align}
f_{\mathsf{lora}}(\bx) = \bW \bx + s \bB \bA \bx
\end{align}

For some low-rank matrices $\bB \in \mathbb{R}^{m\times r}$, $\bA \in \mathbb{R}^{r \times n}$ and a fixed scalar $s \in \mathbb{R}$, where the rank $r$ is often chosen such that $r \ll \min(m, n)$. 

Although the forward pass incurs an extra computational overhead, the significance of LoRA parameterization pertains to the optimizer memory footprint. Optimizers such as AdamW~\citep{kingma2014adam,loshchilov2017decoupled} typically maintain two states for each parameter, resulting in memory consumption that is twice the size of the trainable parameters. In the case of LoRA parameterization, the optimizer memory scales with the combined sizes of $\bA$ and $\bB$. This results in significant memory savings when the memory cost of LoRA $\mathcal{O}(r(m+n))$ is less than the memory cost of the model $\mathcal{O}(mn)$.
Moreover, QLoRA~\citep{dettmers2023qlora} achieves further memory savings by storing $\bW$ in low-precision~(4bit) while keeping the trainable parameters $\bA$ and $\bB$ in higher-precision~(16bit). These works have catalyzed the development of several repositories~\citep{wang2023alpacalora,dettmers2023qlora,dettmers2023bitsandbytes,huggingface2023peft}, enabling finetuning of models with billions of parameters on low-memory devices.

\section{Method}
\label{sec:methods}
To understand the conditions required to pre-train a model with LoRA, we first identify a specific scenario where standard training performance can be recovered using LoRA. This serves as a guide for developing our algorithm that retains LoRA's memory efficiency.

Although low-rank adapters (LoRAs) have proven to be an effective finetuning method, they have apparent limitations when pre-training. As evidenced in Figure~\ref{fig:full_model_vs_lora}, models parameterized with LoRA demonstrate inferior performance compared to models trained using standard optimization. This performance gap isn't surprising as it can be attributed to the inherent rank constraint in LoRA. Specifically, for parameter $\bW \in \mathbb{R}^{m \times n}$, LoRA is fundamentally incapable of recovering weights that exceed the rank $r < \min(m, n)$. Of course, there are exceptions in which, by happenstance, a solution exists within a low-rank proximity of the initialization. However, in~\app{app:vit-rank-dynamics}, we observed the rank of the gradient tends to increase throughout training, hinting at the necessity for high-rank updates.

\begin{figure}[t!]
\centering
    \includegraphics[width=\linewidth]{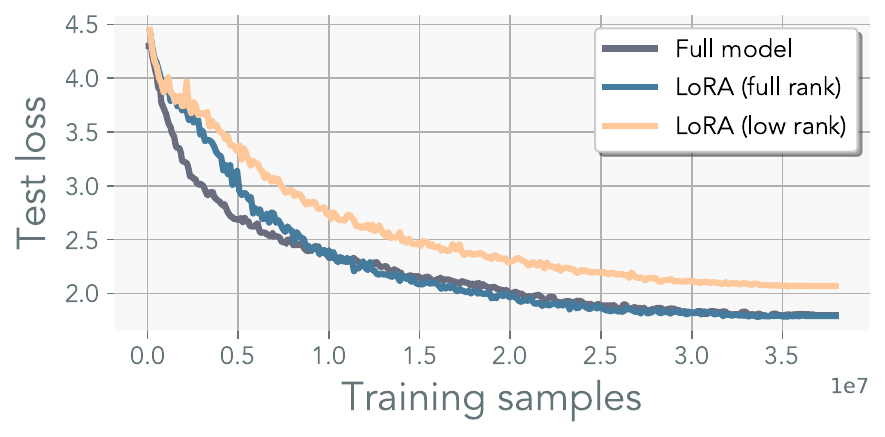}
    \caption{\small\textbf{Increasing the rank of LoRA can recover the standard training performance:} ViT-S trained on ImageNet100 using with and without LoRA. Low-rank LoRA uses rank $r=64$, and full-rank LoRA uses rank $r=\min(m, n)$ set to the dimension of the original weight $\bW \in \mathbb{R}^{m \times n}$. Increasing $r$ suffices to match standard training performance.}
    \label{fig:full_model_vs_lora}
    \vspace{0.1in}
\end{figure}

\subsection{Motivation: Multi-head merging perspective}
\label{sec:motivation}
This section provides intuition on why LoRA heads in parallel can achieve the performance of standard pre-training.

As demonstrated in \fig{fig:full_model_vs_lora}, elevating the rank $r$ of the LoRA to be the same as the rank $\min(m, n)$ of the weight matrix $\bW\in\mathbb{R}^{m\times n}$ is sufficient to replicate standard pre-training performance, albeit with different inherent dynamics as detailed in \app{app:lora_effective_update}. However, such an approach compromises the memory efficiency of low-rank adapters. 

Therefore, we investigate the possibility of arriving at an equivalent performance by leveraging multiple low-rank adapters in \textit{parallel}. Our motivation leverages the trivial idea of linearity of these adapters to induce parallelization.

Given a matrix of the form $\bB\bA \in \mathbb{R}^{d_1 \times d_2}$ with $\bB \in \mathbb{R}^{d_1 \times d}$ and $\bA \in \mathbb{R}^{d \times d_2}$, it is possible to represent the product as the sum of two lower-rank matrices: $\bB_1 \bA_1 + \bB_2 \bA_2$. To demonstrate this, let $\bb_{i}$ and $\ba_{i}$ be the column vectors of $\bB$ and $\bA$ respectively. One can then construct $\bB_1 = [ \bb_{1}, \dots, \bb_{[d/2]} ]$ , $\bB_2 = [ \bb_{[d/2]}, \dots, \bb_{d}]$, and $\bA_1 = [ \ba_{1}^T, \dots, \ba_{[d/2]}^T]$ , $\bA_2 = [ \ba_{[d/2]}^T, \dots, \ba_{d}^T ]$. This decomposition allows for the approximation of high-rank matrices through a linear combination of lower-rank matrices. The same conclusion can be reached by beginning with a linear combination of rank-1 matrices. This forms the basis for a novel multi-head LoRA parameterization, which we will use as one of the baselines to compare with our final method.

\xpar{Multi-head LoRA (MHLoRA)} Given a matrix $\bW \in \mathbb{R}^{m\times n}$, and constant $N$, multi-head LoRA parameterizes the weights as a linear combination of $N$ low-rank matrices $\bB_n$ and $\bA_n$:
\begin{align}
f_{\mathsf{mhlora}}(\bx) = \bW \bx + \frac{s}{N} \sum_{n=1}^{N} \bB_n \bA_n \bx
\end{align}

Multi-head LoRA reparameterizes the full-rank weights into a linear combination of low-rank weights. 

Now, we will point out a trivial observation that a single parallel LoRA head can approximate the trajectory of a single step of the multi-head LoRA, provided that the parallel LoRA heads are periodically merged into the full weights. 

Using the same rank $r$ for all the LoRA parameters, the dynamics of a single parallel LoRA head (denoted with $\hat\cdot$~) is equivalent to multi-head LoRA:
\begin{align}
\argmin_{\bB_n \bA_n} \mathcal{L} &\left( \bW + \frac{s}{N} \sum_{n=1}^{N} \bB_n \bA_n \right) \nonumber \\
&= \argmin_{\hat \bB_n, \hat \bA_n} \mathcal{L} \left( \hat \bW + \frac{s}{N}  \hat \bB_n \hat\bA_n \right)
\end{align}

When either $\sum_{n=1}^{N} \bB_n \bA_n $ is equal to $\hat \bB_n \hat \bA_n$, or when $\hat \bW = \bW + \frac{s}{N} \sum_{j \neq n}^N \bB_j  \bA_j$; here we used a shorthand notation to indicate that sum is over all the LoRA parameters except for index $n$. We assume the parameters on both sides of the equation are initialized to be the same: $\bA_n = \hat\bA_n$ and $\bB_n = \hat\bB_n \forall n$. The first scenario is rank deficient, which we know is unable to recover the original model performance. The latter case necessitates that $\hat\bW$ accumulates all the information of the LoRA parameters at every iteration. Hence, if we can apply a merge operator at every iteration, we can recover the exact update.

This rather simple observation implies that one can recover the exact gradient updates of the multi-head LoRA parameterized model, which we observed to match pre-training performance across a wide range of tasks~(see~\app{app:more-results}). Moreover, in a distributed setting, only the LoRA parameters/gradients have to be communicated across devices, which is often a fraction of the original model size, making it a good candidate where interconnect speed between computing nodes is limited.

\begin{figure*}[t!]
    \centering
    \includegraphics[width=1.0\linewidth]{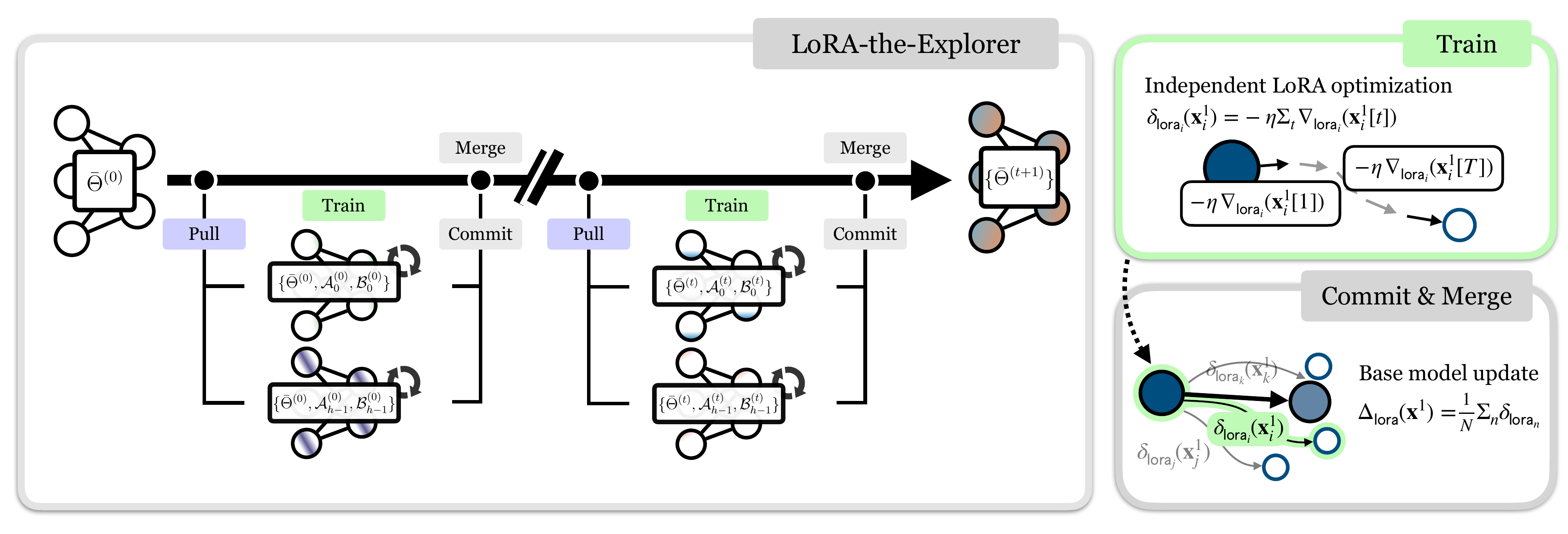}
    \caption{\small \textbf{LTE diagram}:
    Our method is decomposed into 3 steps. (\textbf{1}) We parameterize the model with multiple LoRA heads and train them independently for $T$ iterations using different mini-batches sampled from the same (homogeneous) distribution. This results in overall update of $\delta_{\mathsf{lora}_n}(\bx) = - 
    \eta \sum_t \nabla_{\mathsf{lora}_n}(\bx[t])$ (\textbf{2}). Next, we accumulate the individual LoRA updates by averaging the heads $\Delta_{\mathsf{lora}}(\bx) = \frac{1}{N}\sum_n \delta_{\mathsf{lora}_n}(\bx)$. (\textbf{3}) The update is applied to the main weights, and the LoRA parameter $\bB$ is reset. The optimization repeats with the new LoRA parameters. 
    LTE resembles the distributed model development paradigm first proposed by~\citet{kandpal2023git}.
    }
    \label{fig:lte}
\end{figure*}

\subsection{LoRA soup: delayed LoRA merging}

To further reduce the communication cost of LTE, we extend and combine the ideas of local updates~\cite{mcmahan2017communication} and model-averaging~\cite{wortsman2022model,yadav2023ties-merging,ilharco2022editing}. 
Instead of merging every iteration, we allow the LoRA parameters to train independently of each other for a longer period before the merge operator. This is equivalent to using stale estimates of the LoRA parameters $\hat \bW = \bW + \frac{s}{N} \sum_{j \neq n}^N \bB_j'  \bA_j'$ with $'$ indicating a stale estimate of the parameters. 

\begin{figure}[t!]
    \centering
    \subfigure[Gradient deviation]{\label{fig:lora-div-a}\includegraphics[width=0.48\linewidth]{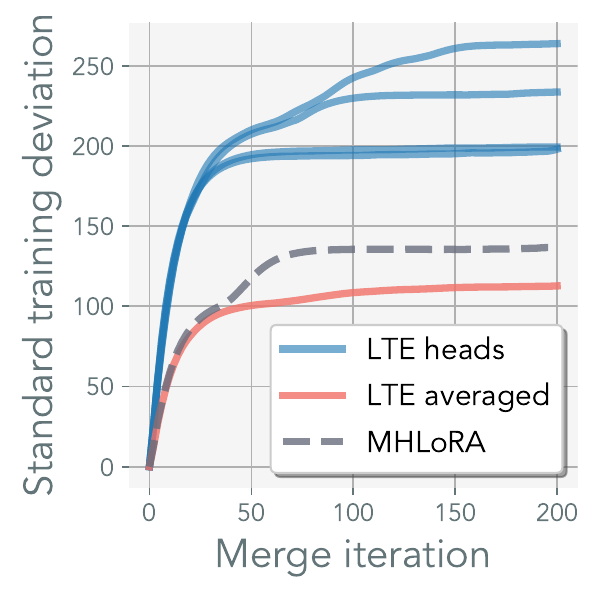}}
    \hfill
    \subfigure[Trajectory projection]{\label{fig:lora-div-b}\includegraphics[width=0.48\linewidth]{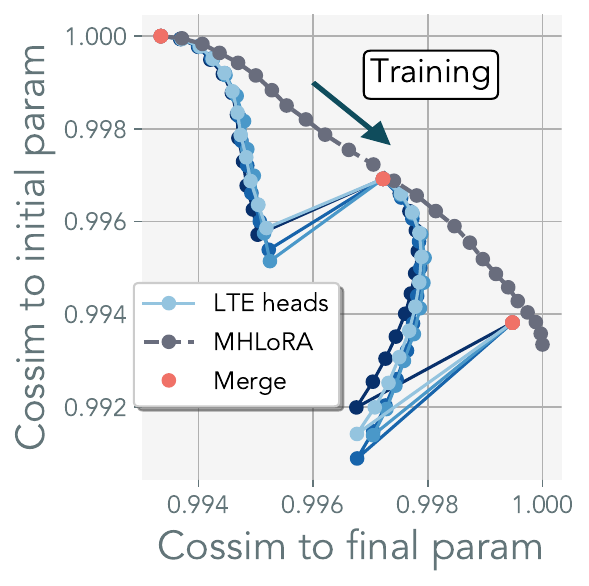}}
    \caption{\small \textbf{Effects of merging LoRA heads.} \textbf{Left:} We measure $l_2$-norm deviation of the effective weights of multi-head LoRA (MHLoRA) and LoRA-the-explorer (LTE, our method) from the weights of standard training using ViT-S. We use $4$ heads for both MHLoRA and LTE using the same initialization, and we measure the norm of \texttt{encoder-layer-3}. We also plot the individual LoRA heads of LTE. These heads deviate more from standard training, but their average closely follows that of MHLoRA. Depending on the merge iteration (x-axis), the estimation gap of using stale estimates is roughly the difference between the MHLoRA and LTE averaged. The later the merge happens, the more LTE deviates from MHLoRA. \textbf{Right:} We project the dynamics of MHLoRA and LTE onto the parameters of MHLoRA. The y-axis is the initial parameters, and the x-axis is after training for $25$ iterations. The projection is computed by computing the cosine similarity on the vectorized weights and creating an arc from (0, 1) to (1, 0). We set merge iteration to $12$ and visualize how the LTE trajectory follows the arc of MHLoRA. }
    \label{fig:lora-divergence}
\end{figure}

Merging every iteration ensures that the representation will not diverge from the intended update. While using stale estimates relaxes this equivalence, we observe that it can still match the standard training performance as shown in~\tbl{table:imagenet100-ablation}. 
Nevertheless, as the estimate becomes inaccurate, the optimization trajectory does indeed diverge from the optimization path of multi-head LoRA. We quantify this divergence in~\fig{fig:lora-divergence}. The divergence does not imply that the model won't optimize; rather, it suggests that the optimization trajectory will deviate from that of the multi-head LoRA. 
In this work, we opt for simple averaging and leave more sophisticated merging such as those used in~\cite {karimireddy2020scaffold,matena2022merging,yadav2023ties-merging} for future works.

\subsection{LoRA-the-Explorer: parallel low-rank updates}
\label{sec:lte}

Our algorithm is designed with two primary considerations: (1) achieving an informative update $\Delta \bW$ that does not require materialization of the full parameter size during training, and (2) parameterizing $\bW$ such that it can be stored in low-precision and communicated efficiently. The latter can be achieved by using quantized weights and keeping a high-precision copy of $\bW$.

We propose LoRA-the-Explorer (LTE), an optimization algorithm that approximates full-rank updates with parallel low-rank updates. The algorithm creates $N$-different LoRA parameters for each linear layer at initialization. Each worker is assigned the LoRA parameter and creates a local optimizer. Next, the data is independently sampled from the same distribution $\bx = \{ \bx_1, \dots \bx_N \}$. For each LoRA head $n$, the parameters are optimized with respect to its own data partition for $T$ iterations resulting in an update $\delta_{\mathsf{lora_n}} = - \eta \sum_{t=1}^T \nabla_{\mathsf{lora}_n} \bx_i[t]$. 
We do not synchronize the optimizer state across workers.
After the optimization, the resulting LoRA parameters are synchronized to compute the final update for the main weight $\Delta_{\mathsf{lora}}(\bx) = \frac{1}{N} \sum_{n=1}^N \delta_{n}$. In the next training cycle, the LoRA parameters are trained with the updated weights $\bW$. Here, the LoRA parameters can be either be re-initialized or the same parameters can be used with the correction term~(see~\app{app:exact}). Since we do not train directly on the main parameter $\bW$, we can use the quantized parameter $q(\bW)$ instead. Where one can either keep the high-precision weight only in the master node or offload it from the device during training. This reduces not only the memory footprint of each worker but also the transmission overhead. A pseudo-code is provided in~\alg{alg:lte} and an illustration in~\fig{fig:lte}.

\subsection{Implementation details}
\label{sec:implementation}

We discuss a few of the implementation details we found necessary for improving the convergence speed and the performance of our method. Full training details and the supporting experiments can be found in~\app{app:appendix}. 

\xpar{Not resetting matrix $\bA$ and optimizer states}
We investigate whether the matrices $\bA_n$ would converge to the same sub-space during training. If so, it would necessitate resetting of matrices $\bA_n$ or the use of a regularizer. 
In~\fig{fig:lora-alignment}, we did not observe this to be the case. We observed the orthogonality of $\bA$ to remain consistent throughout training, and we found it to perform better without resets. We posit that re-learning matrix $\bA$ and re-accumulating the optimizer state ends up wasting optimization steps. The comparison figures can be found in~\app{app:ablation} and a more detailed discussion in~\sect{sec:lora-alignment}.

\xpar{Scaling up $s$ and lowering learning rate $\eta$}

It is a common misconception that scaling $s$ has the same effect as tuning the learning rate $\eta$. During our experimentation, we were unable to yield comparable performance when using the standard value of $s$ (in the range of $1\sim 4$). Instead, we found using a large value of $s$ and a slightly lower learning rate $\eta$ to work the best. The standard practice is to set the scaling proportionately to the rank of the LoRA $s = \alpha / r$. This is done to automatically adjust for the rank~\citep{hu2021lora}. We use $\alpha=4096$ ($s=64$) and a learning rate of $\eta=2\cdot10^{-4}$. It is worth noting that the learning rate does not scale linearly with $s$, and the scalar only affects the forward computation (\app{app:s_vs_lr}).
The scalar $s$ modifies the contribution of the LoRA parameters in the forward pass, which has a non-trivial implication on the effective gradient. Moreover, in~\app{app:lora_effective_update}, we provide the effective update rule of LoRA updates and observe the emergence of the term $s$ that scales quadratically with the alignment of $\bB$ and $\bA$. Since $s$ is large relative to the learning rate, it may have a non-negligible effect on the dynamics.

\newcommand{\algrule}[1][.2pt]{\par\vskip.5\baselineskip\hrule height #1\par\vskip.5\baselineskip}

\begin{figure}[t!]
\centering
 \scalebox{0.95}{
\begin{minipage}{1.0\linewidth}
  \begin{algorithm}[H]
  \DontPrintSemicolon
    \SetCustomAlgoRuledWidth{0.55\textwidth}  
    \caption{LoRA-the-Explorer (LTE)}\label{alg:lte}
    \KwInput{Dataset $\mathcal{D}_{\mathsf{train}}$,
        model $\mathcal{F}$, loss function $\mathcal{L}$ \\
        \qquad \quad ~parameters $\Theta = \{ \bW_0, \dots, \bW_L\} $, \\
        \qquad \quad ~merge scalar $s$, num workers $N$, merge iter $T$}
        \algrule
     \While{not converged}{
    (\textit{Optional}) quantize $\Theta$. Keep high-precision copy\;\\
      \textbf{\color{teal} (in parallel) }\For{each worker $n$}{
          \If{LoRA not initialized}{
              $\mathcal{B}_n, \mathcal{A}_n \leftarrow \mathbf{lora\_parameterize}(\mathcal{F})$\;
          }
          \Else{
              (\textit{Optional}) reset parameters $\mathcal{B}_n$ to zero\;
          }
         Optimize $\mathcal{B}_n, \mathcal{A}_n$ for $T$ iterations by minimizing $\mathbb{E}_{\bx, \by \sim \mathcal{D}_{\mathsf{train}}} \left[ \mathcal{L}(\mathcal{F}(\bx), \by) \right]$ \;
      } \vspace{0.1in}
        \; \# Synchronize by communicating LoRA parameters.\;
        \For{each worker $n$}{
            \For{$\bB_n,\bA_n$ in $\mathcal{B}_n, \mathcal{A}_n$}{
            Merge LoRA params $\bW_n \leftarrow \bW_n + \frac{s}{n} \bB_n \bA_n$\;
            }
      }
     }
    \vspace{0.1in}
  \end{algorithm}
\end{minipage}
}
\caption{\small LTE pseudocode. Not resetting $\mathcal{B}_n$ requires a correction term to LTE. See~\app{app:exact}.}
\end{figure}

\xpar{Significance of Initialization Strategies}
\label{sec:init_matters}

Initialization of LoRA plays a pivotal role in pre-training. Kaiming initialization used in the original work~\citep{hu2021lora} -- are not well-suited for rectangular matrices as discussed in~\citep{bernstein2023agd,yang2020feature}. Given that LoRA parameterization often leads to wide matrices, alternative methods from~\citep{bernstein2023agd} and~\citep{glorot2010understanding} resulted in better empirical performance.

We use the initialization scheme prescribed in~\citep{bernstein2023agd} that utilizes a semi-orthogonal matrix scaled by $\sqrt{d_{out}/d_{in}}$. Note that these methods were originally designed for standard feed-forward models. Whereas LoRA operates under the assumption that matrix $\bB$ is zero-initialized with a residual connection. This aspect warrants further study for exact gain calculations. 
Our ablation studies, in~\app{app:ablation}, indicate the best performance with~\cite{bernstein2023agd}, with Kaiming and Xavier initializations performing similar. In ImageNet-1k, we found the performance gap to be more evident.

\section{Experiments}
\label{sec:experiments}
We follow standard training protocols, and all implementation details and training hyper-parameters can be found in~\app{app:train-details}.

\textbf{Disclaimer:} During the preparation of the code release, we found that the transformer experiments used a scaling factor of $1/\sqrt{d_{out}}$ instead of the standard scaling $1/\sqrt{d_{out}/n_{attn}}$~\cite{vaswani2017attention}. Hence, using LTE with the standard scaling would require a different set of hyper-parameters than the ones reported in~\app{app:train-details}. We will revise the hyper-parameters in the next revision.

\begin{figure}[t!]
\centering
    \includegraphics[width=\linewidth]{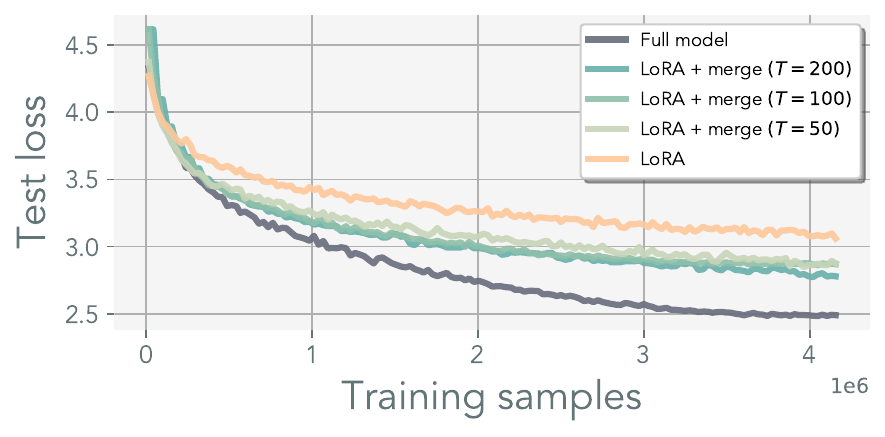}
    \caption{\small\textbf{Sequential merging of LoRA cannot recover performance.} ViT-S trained on ImageNet100. Merging and resetting the LoRA parameters achieves better performance than single-head LoRA pre-training but still cannot recover the standard full-model pre-training. Note that LoRA + merge is akin to concurrent work of ReLoRA~\citep{lialin2023stack}.
    }
    \label{fig:full_model_vs_mergelora}
\end{figure}

\subsection{Iterative LoRA Merging}
\label{sec:merging}
In~\sect{sec:motivation}, we motivated that iteratively merging LoRA parameters is a key component in accurately recovering the full-rank representation of the model. 
As a sanity check, in~\app{app:merge-least-squares}, we assess the effectiveness of merging a single LoRA head in the context of linear networks trained on synthetic least-squares regression datasets. The underlying rank of the optimal solution, $\bW^*$, is controlled, and datasets are generated as $\bY = \bX (\bW^*)^{\mathsf{T}}$. Each $\bx \in \bX$ follows a normal distribution, $\bx \sim \mathcal{N}(\mathbf{0}, \mathbf{I})$.
Figure \ref{fig:merge_ls} evaluates the model's rank recovery across varying merge iteration $T$. Dimension of the weights $\bW$ are set to $m=n=32$. 
Without merging, the model performance plateaus rapidly on full-rank $\bW^*$. In contrast, iterative merging recovers the ground truth solution with the rate increasing with higher merge frequency. 

Further tests in~\fig{fig:full_model_vs_mergelora} using ViT-S~\citep{dosovitskiy2020image} with a patch-size of 32 on the ImageNet100 dataset~\citep{tian2020contrastive} (a subset of ImageNet~\citep{russakovsky2015imagenet}) confirm that merging of a single LoRA head outperforms standalone LoRA parameter training. However, frequent merging delays convergence, likely due to LoRA parameter re-initialization and momentum state inconsistencies. Additionally, the performance does not match that of fully trained models, indicating potential local minima when training with rank-deficient representations. 

We find that the merge iteration of $T=10$ is still stable when using batch size $4096$. With higher $T$ values, additional training may be required to achieve comparable performance~\citep{stich2018local,wang2021cooperative,yu2019parallel}. Our initial efforts to improve merging using methods such as~\citep{yadav2023ties-merging} did yield better results. Nonetheless, we believe with increased merge iteration, smarter merging techniques may be necessary. Existing literature in federated learning and linear-mode connectivity/model-averaging may provide insights into designing better merging criteria.

\begin{figure*}[t!]
    \centering
    \includegraphics[width=1.0\linewidth]{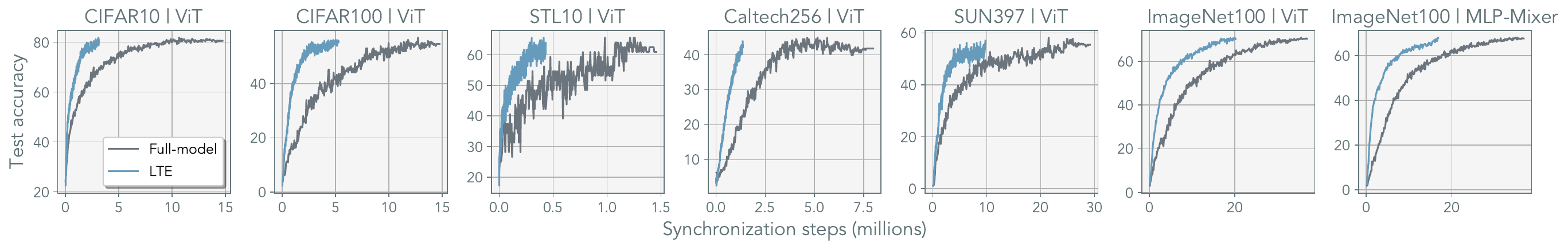}\\
    \includegraphics[width=1.0\linewidth]{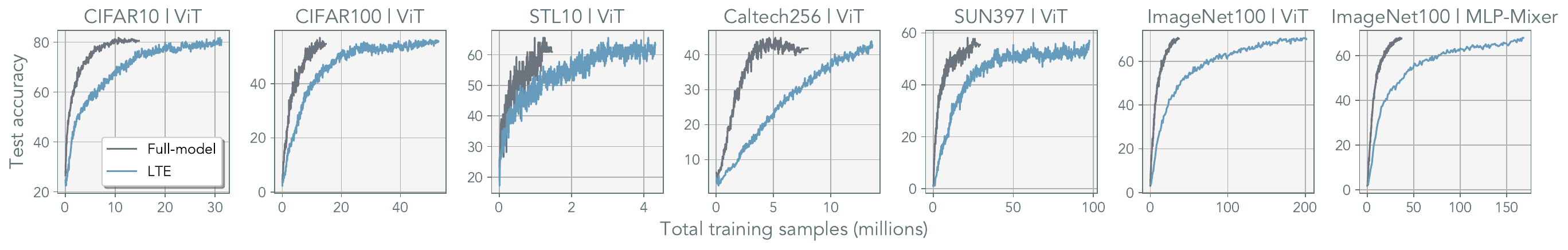}
    \caption{\small \textbf{Experiments on various vision tasks}: We apply our method to a range of vision tasks, including CIFAR10, CIFAR100~\cite{krizhevsky2009learning}, STL10~\cite{coates2011analysis}, Caltech256~\cite{griffin2007caltech}, SUN397~\cite{xiao2010sun}, and ImageNet100~\cite{tian2020contrastive}. Details of these datasets are listed in~\app{app:datasets}. Additionally, we incorporated the MLP-mixer and observed consistent results. The figure is presented with two different x-axes. On the top, we plot the total number of synchronization steps, and on the bottom, we measure the total number of training samples observed across all devices. All LTE experiments are conducted on ViT-S with a merge iteration of $T=10$ and 32 LoRA heads with a rank of $r=64$. In~\app{app:more-results}, we provide additional results for $T=1$, LLM experiments, and LTE trained on ViT and MLP-mixer at various scales.
    }
    \label{fig:results}
\end{figure*}

To further test the generalizability of our method, we conducted a suite of experiments on various vision tasks in~\fig{fig:results}. Moreover, we test our method on MLP-Mixer~\cite{tolstikhin2021mlp} to demonstrate its use outside of transformer architecture. We provide additional experiments when using $T=1$ and initial language-modeling results in~\app{app:more-results}.

\subsection{LoRA parameter alignment}
\label{sec:lora-alignment}
\begin{figure}[t!]
\centering

    \includegraphics[width=1.0\linewidth]{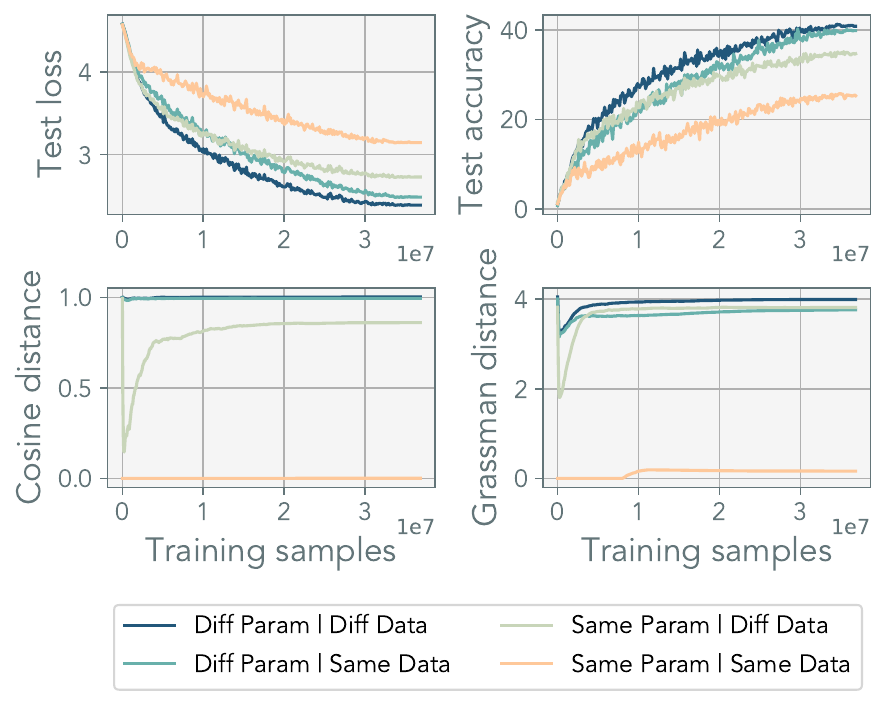}

    \caption{\small\textbf{LoRA alignment}: Alignment of LTE heads when varying parameters and data. ``Diff\;Param'' uses random initialization across each head, and ``Diff\;Data'' uses different mini-batches. The similarity is computed on the first epoch of ImageNet100 on ViT-S. We use LTE of $r$=$8$ with $4$ LoRA heads. Pair-wise similarity is averaged across all linear layers. The performance of the model correlates with orthogonality between LoRA heads.
    }
    \label{fig:lora-alignment}
\end{figure}

The efficacy of our optimization algorithm hinges on the ability of individual heads to explore distinct subspaces within the parameter space. We examine the extent to which data and initial parameters influence intra-head similarity throughout training. In~\fig{fig:lora-alignment}, we compute the average cosine similarity and Grassman distance~(see~\app{app:grass}) between the heads $\bB_n\bA_n$. These tests were conducted with data samples drawn from the same distribution, and each set of LoRA parameters was exposed to a different set of samples. 

Our results confirm that LoRA heads do not converge to the same representation. We find using different initializations across LoRA heads yields the greatest orthogonality. This orthogonality is further increased when different mini-batches are used for each head. Importantly, the degree of alignment among LoRA heads remains stable post-initialization and does not collapse into the same representation. In~\fig{fig:lora-alignment}, we find that lower similarity corresponds well with model performance, where using different parameters and mini-batches significantly outperforms other configurations. 

\begin{figure}[t!]
  \centering
  \begin{minipage}[b]{\linewidth}
    \centering
    \scalebox{0.7}{
    \begin{tabular}{l|cccc|ccc|cc} 
    \toprule
    & LTE & Heads & Rank & Merge & Test loss $\downarrow$ & Test acc $\uparrow$ \\
    \midrule
    & - & - & - & - & 1.78 & 71.97 \\
    \cmidrule(lr){1-7}
    {\multirow{6}{*}{\rotatebox[origin=c]{90}{$\mathsf{head}$}}}
    & \checkmark & 2 & 8 & 10 & 2.31 & 54.68\\
    & \checkmark & 8 & 8 & 10 & 2.35 & 54.88\\
    & \checkmark & 32 & 8 & 10 & 2.26 & 58.21 \\
    & \checkmark & 2 & 64 & 10 & 1.86 & 69.82 \\
    & \checkmark & 8 & 64 & 10 & 1.84 & 71.97  \\
    & \checkmark & 32 & 64 & 10 & 1.79 & 73.73 \\
    \cmidrule(lr){1-7}
    {\multirow{5}{*}{\rotatebox[origin=c]{90}{$\mathsf{rank}$}}}
    & \checkmark & 32 & 8 & 10 & 2.66 & 44.00\\
    & \checkmark & 32 & 16 & 10 & 2.12 & 60.35\\
    & \checkmark & 32 & 32 & 10 & 1.81 & 71.20 \\
    & \checkmark & 32 & 64 & 10 & 1.79 & 73.73 \\
    & \checkmark & 32 & 128 & 10 & 1.78 & 73.54 \\
    \cmidrule(lr){1-7}
    {\multirow{5}{*}{\rotatebox[origin=c]{90}{$\mathsf{merge}$}}}
    & \checkmark & 32 & 64 & 5  & 1.87 & 70.67\\
    & \checkmark & 32 & 64 & 10 & 1.79 & 73.73 \\
    & \checkmark & 32 & 64 & 20 & 1.97 & 66.80\\
    & \checkmark & 32 & 64 & 50 & 1.99 & 65.43 \\
    & \checkmark & 32 & 64 & 100 & 2.10 & 61.04 \\
    \bottomrule
    \end{tabular}
    }
    \captionof{table}{\small \textbf{LTE ablation for ViT-S trained ImageNet100:} For \textit{fixed cumulative training epoch} of $1200$, we vary the number of heads, rank, and merge iteration of our method. More heads require longer cumulative training samples to converge; see~\fig {fig:lte_sync_steps}.}
    \label{table:imagenet100-ablation}
  \end{minipage}%
  \vspace{0.3in}
  \begin{minipage}[b]{0.45\textwidth}
    \centering
    \includegraphics[width=\linewidth]{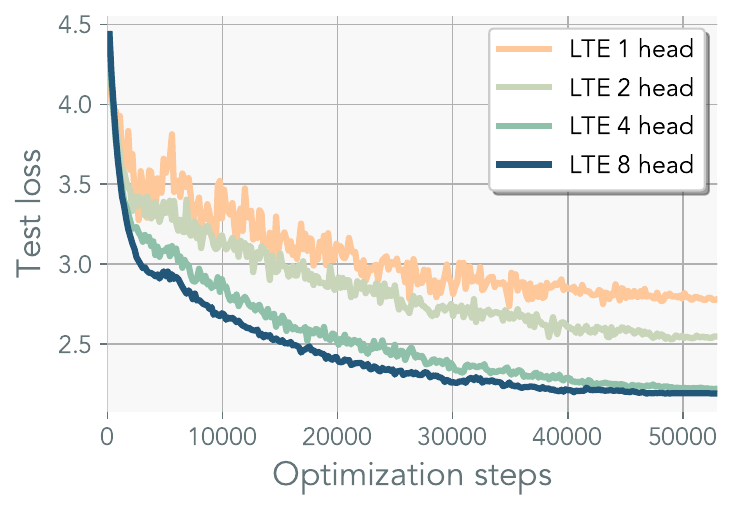}
    \caption{\small\textbf{LTE with same batch-size per head:} ViT-S trained on ImageNet100. We use the same batch-size for each LoRA head with rank $r=8$. In contrast to other figures, we plot the loss in optimization steps. LTE with more heads converge to a better solution but require longer training samples to converge.}
    \label{fig:lte_sync_steps}
  \end{minipage}
\end{figure}

\subsection{Ablation study: the effect of LoRA heads, rank, and merge iteration}
\label{sec:ablation}
We systematically evaluate the effects of varying the number of LoRA heads, rank, and merge iteration on model performance for ImageNet100 in~\tbl{table:imagenet100-ablation}. Our findings indicate a monotonic improvement in performance with an increased number of heads and ranks. Conversely, extending the merge iteration negatively impacts performance. As in the case of least-squares regression, we found excessive merging to hurt model accuracy. With a large enough rank and head, we found the model to converge to better test accuracy, even if the test loss was similar. We hypothesize that the averaging of the LoRA heads has a regularization effect similar to those seen in model ensembling. 

We use ViT-S as the primary architecture for analysis, which has a hidden dimension of $384$ and an MLP dimension of $1536$. We find that setting the product of the number of heads and the rank of the LoRA larger than the largest dimension of the model serves as a good proxy for configuring LTE. For example, using $32$ heads with $r=64$ results in $2048 > 1536$. However, when it comes to increasing the number of heads rather than rank, we noticed longer training iterations were required to achieve comparable performance. In the preceding section, we discuss a potential cause of the slowdown in convergence. 

\subsection{Gradient noise with parallel updates}
In our ablation study, we utilized a fixed cumulative batch size of $4096$ and a training epoch of $1200$. Each LoRA head received a reduced batch size of $\frac{4096}{\text{heads}}$. Our findings indicate that scaling the rank exerts a greater impact than increasing the number of heads. Due to the proportional scaling of gradient noise with smaller mini-batches~\citep{mccandlish2018empirical,shallue2019measuring,smith2017bayesian}, we hypothesize that gradient noise is the primary factor contributing to slower convergence, in addition to the use of stale parameter estimates. To validate this hypothesis, we employed the same mini-batch size across all heads in~\fig{fig:lte_sync_steps}, using a reduced rank of $r=8$. When we adjusted the batch size in proportion to the number of heads and measured it with respect to the optimization steps, the impact of varying the number of heads became more pronounced. While increasing the number of heads necessitates more sequential FLOPs, it offers efficient parallelization. Furthermore, using a larger batch size for gradient estimation may prove beneficial in distributed training, as it increases the computational workload on local devices. Careful optimization of the effective batch size to maximize the signal-to-noise ratio may be crucial for achieving maximum FLOP efficiency.

\begin{figure}[t!]
    \centering
    \includegraphics[width=1.0\linewidth]{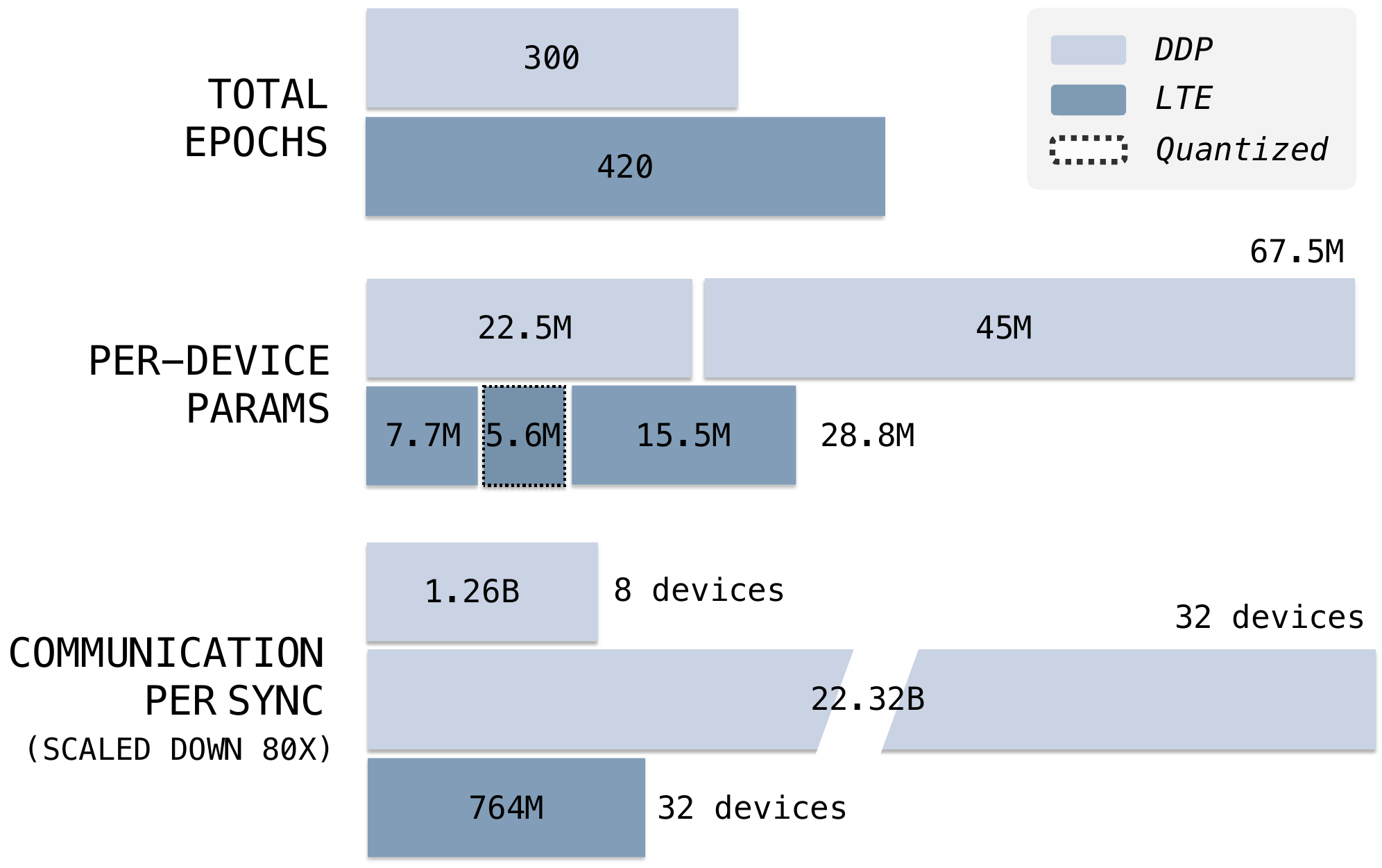}
    \caption{\small \textbf{ImageNet compute analysis}: We break down the computational cost for training ViT-S on ImageNet1k. We compare against distributed data-parallel with $8$ devices. LTE requires $40\%$ longer to achieve the same performance of $68\%$ top-1 accuracy on $1000$-way classification. We used $32$ LoRA heads with $r=64$. Our method requires fewer trainable parameters per device. This, in turn, enables the fitting of larger models in low-memory devices. With a smaller memory footprint and infrequent communication, our method requires lower communication bandwidth. Further discussion is in~\sect{sec:imagenet1k}. 
    }
    \label{fig:lte-analysis}
\end{figure}

\subsection{Performance Scaling on ImageNet-1K}
\label{sec:imagenet1k}
We scaled up our method to ImageNet-1K. We followed the training protocols detailed in~\app{app:train-details}. In accordance with our initial hypothesis on gradient noise, we doubled the batch size to $8192$ (see~\app{app:batch_size_imagenet1k}). Since using different mini-batches was crucial early in training, we did not alter the way mini-batches were sampled. Scheduling the randomness for the mini-batches is an option we have not yet explored.

In the initial training phase, we observed that LTE outperformed standard training. However, as training approached completion, standard training overtook LTE, necessitating additional iterations for LTE to achieve comparable performance. Standard training appeared to benefit more from a lower learning rate compared to LTE. For ViT-S, the model took $40\%$ more training samples to converge to the same top-1 accuracy of $68\%$ (see~\app{app:imagenet1k_test_curve}). 

The primary focus of our work was to investigate whether it is possible to train deep networks with parallel low-rank adapters; hence, we did not aim to maximize efficiency. However, we do provide a hypothetical computation analysis for future scaling efforts. Let the model size be denoted by $M_{\mathsf{ddp}} = M$, and $M_{\mathsf{lte}}$ for LTE, and the respective number of devices for each method be denoted with $N_{\mathsf{ddp}}$, and $N_{\mathsf{lte}}$. With quantization, each LTE device would require a memory footprint of $qM + M_{\mathsf{lte}}$. With the base model operating in 16-bit precision, using 4-bit quantization results in $q = 0.25$. With AdamW, DDP necessitates an additional $2M$ parameters, making the total memory footprint $3M$ per device. For LTE, the total memory footprint per device is $qM + 3M_{\mathsf{lte}}$. Assuming the training is parameter-bound by the main weights $r \ll \min(m, n)$, LTE can leverage GPUs that are roughly $1/3$ the size required for DDP.
It is worth noting that LTE requires $40\%$ more data to train and a slowdown of $20\%$ per iteration when using quantization methods such as QLoRA. If the cost of low-memory devices is lower, these slowdowns may be negligible compared to the speed-up achieved through parallelization. On average, each LTE device observes $1/3$ less data than a device in DDP. With improvements in our method and future advances in quantization, we believe this gap will be reduced. The compute analysis for ViT-S on ImageNet1k is illustrated in~\fig{fig:lte-analysis}.

Communication also presents bottlenecks when training models across nodes. For a single node, one can interleave the communication of gradients asynchronously with the backward pass~\citep{li2020pytorch}. In multi-node systems, the communication scales with the size of the trained parameters and is bottlenecked at interconnect speed, especially when high-throughput communication hardware, such as InfiniBand, is not utilized. 
Using standard all-reduce, the gradient is shared between each device for a total communication of $N_{\mathsf{ddp}} (N_{\mathsf{ddp}}-1)M$. For LTE we communicate every $T$ iteration hence we have $\frac{1}{T} N_{\mathsf{lte}} (N_{\mathsf{lte}}-1)M$. To maximize the efficacy of LTE, an alternative approach is to use a parameter server for 1-and-broadcast communication. Here, gradients are sent to the main parameter server and averaged. The accumulated updates are broadcast back to other nodes. DDP with a parameter server would use $2 (N_{\mathsf{ddp}}-1)M$ and LTE would use $\frac{1}{T} ((N_{\mathsf{lte}}-1)M_{\mathsf{lte}} + (N_{\mathsf{lte}}-1)qM)$.
Moreover, LTE can leverage lower-bandwidth communication since the parameters shared between devices are strictly smaller by a factor of $M_{\mathsf{ddp}}/ M_{\mathsf{lte}}$. 

\section{Related works}
\label{sec:related}

\paragraph{Training with adapters}

The use of LoRA has received considerable attention in recent literature. While we focused on using LoRA to pre-train models from scratch, prior works have mainly focused on finetuning~\cite{chavan2023one, zhang2023adaptive}. The flexibility of LoRA has had a broad reach in different aspects of model training, from reducing computational requirements~\citep{zhang2023lora} to offsetting part of the pre-training computation~\citep{lialin2023stack}. 

LoRA is just one type of adapter, and various forms of adapters have been explored. Parallel additive adapters~\citep{zhang2021tip} augment existing models with additional parameters. Linear adapters have also been common, either in the form of residual connections~\citep{cai2020tinytl} or affine parameters in normalization layers~\citep{bettelli2006reciprocal,mudrakarta2018k}. 

Adapters have been applied to numerous tasks: natural language processing~\citep{houlsby2019parameter,stickland2019bert}, video~\citep{yang2023probabilistic,xing2023simda}, computer vision~\citep{sax2019side,zhang2023adding,chen2023vision}, incremental learning~\citep{rosenfeld2018incremental}, domain adaptation~\citep{rebuffi2018efficient}, and vision-language tasks~\citep{gao2023clip,radford2021learning,sung2022vl}, text-to-vision generative models~\citep{mou2023t2i}, and even perceptual learning~\citep{fu2023dreamsim}. 

\vspace{-0.1in}
\paragraph{Distributed Training and Federated Learning}
Our work shares relevance to both distributed and federated learning paradigms, where the LoRA head in our work can be conceptualized as a distinct computational device. Our work shares many motivations for federated learning, including low-compute devices, high-latency training, privacy, and cross and in-silo learning. See~\citep{mcmahan2017communication,wang2021field} for a comprehensive discussion.

Communication efficiency serves as a cornerstone in both distributed and federated learning. Techniques such as \textit{local steps} have been employed to mitigate communication load~\citep{mcmahan2017communication,lin2018don,povey2014parallel,smith2018cocoa,su2015experiments,zhang2016parallel}. These methods defer the averaging of weights to specific optimization steps, thus reducing the communication cost per iteration. The effectiveness of decentralized training has been studied in~\citep{lian2017can,koloskova2019decentralized,koloskova2020unified, coquelin2022accelerating}.

Traditionally, activation computations have dominated the computational load. However, the advent of gradient checkpointing~\citep{chen2016training} and reversible gradient computation~\citep{gomez2017reversible,mangalam2022reversible} has shifted the training process toward being parameter-bound. Techniques such as gradient or weight compression also seek to reduce the communication burden~\cite {lin2017deep,aji2017sparse,wen2017terngrad}.

Combining models in federated learning is often credited to FedAvg~\citep{mcmahan2017communication}. Numerous studies explore the use of weighted averaging to improve convergence speed~\citep{li2019convergence}. Since then, many works have tried to use probabilistic frameworks to understand and improve merging~\citep{hsu2019measuring,wang2019slowmo,reddi2020adaptive}. The conditions for optimal merging is still an open question, with recent efforts to improve updating with stale parameters~\cite{chen2022sapipe}.

Server momentum and adaptive methods constitute another active area of research where macro synchronization steps are interpreted as ``gradients'', allowing one to prescribe a bi-level optimization scheme~\citep{hsu2019measuring,wang2019slowmo,reddi2020adaptive}. 

Initial efforts have been made to use federated learning with large models. \cite{yuan2022decentralized} examined the cost models for pre-training Language Learning Models (LLMs) in a decentralized configuration. \cite{wang2023cocktailsgd} suggested the utilization of compressed sparse optimization methods for efficient communication. Our work is also closely related to~\cite{douillard2023diloco}, a concurrent work that explores the idea of distributing computing across smaller server farms while maintaining low communication costs. 

\vspace{-0.1in}
\paragraph{Linear mode connectivity and model averaging}
Linear mode connectivity~\citep{garipov2018loss} studies the phenomena of why and when models are smoothly connected~\citep{freeman2016topology,draxler2018essentially,fort2019large}. Under the same initialization, models have been shown to have a linear path with constant energy~\citep{nagarajan2019uniform,frankle2020linear,wortsman2022model}. For models with different initializations, parameter permutations can be solved to align them linearly~\citep{brea2019weight,tatro2020optimizing,entezari2021role,simsek2021geometry}.

Following this line of research, numerous works have also explored model averaging and stitching. Where averaging of large models has shown to improve performance~\citep{wortsman2022model,ainsworth2023git,stoica2023zipit,jordan2022repair,wortsman2022fi}. Model stitching~\citep{lenc2015understanding} has also shown to yield surprising transfer capabilities~\citep{moschella2022relative}. This idea is conceptually related to optimal averaging in convex problems~\citep{scaman2019optimal} and the ``Anna Karenina'' principle where successful models converge to similar solutions~\citep{bansal2021revisiting}. This phenomenon could help explain the success of averaging multiple LoRAs into a single global LoRA~\cite{yi2023fedlora} and averaging MoE MLP layers into a single weight at inference~\cite{wang2022adamix}.

\section{Conclusion}
\label{sec:discussion}
In this work, we investigated the use of low-rank adapters for model pre-training. We introduced LTE, a bi-level optimization method that capitalizes on the memory-efficient properties of LoRA. Although we succeeded in matching performance on moderately sized tasks, several questions remain unresolved. These include: how to accelerate convergence during the final $10\%$ of training; how to dynamically determine the number of ranks or heads required; whether heterogeneous parameterization of LoRA is feasible, where each LoRA head employs a variable rank $r$; and leveraging merging strategies to accompany higher local optimization steps. Our work serves as a proof-of-concept, demonstrating the viability of utilizing low-rank adapters for neural network training from scratch. However, stress tests on larger models are essential for a comprehensive understanding of the method's scalability. Addressing these open questions will be crucial for understanding the limitations of our approach. We anticipate that our work will pave the way for pre-training models in computationally constrained or low-bandwidth environments, where less capable and low-memory devices can collaboratively train a large model, embodying the concept of the ``\textit{wisdom of the crowd.}''

\section{Acknowledgement}
The research was sponsored by the Army Research Office and was accomplished under Grant Number W911NF-23-1-0277. The views and conclusions contained in this document are those of the authors and should not be interpreted as representing the official policies, either expressed or implied, of the Army Research Office or the U.S. Government.
MH was also supported by the ONR MURI grant N00014-22-1-2740 and the MIT-IBM Watson AI Lab. 
JB is supported by the MIT-IBM Watson AI Lab and the Packard Fellowship.
PI is supported by the Packard Fellowship.
BC is supported by the NSF STC award CCF-1231216.
We thank Han Guo, Lucy Chai, Wei-Chiu Ma, Eunice Lee, and Yen-Chen Lin for their feedback.
\bibliography{refs}
\bibliographystyle{icml2024}

\newpage
\appendix
\onecolumn
\appendix

\section{Appendix}
\label{app:appendix}

\subsection{Training details}
\label{app:train-details}

\xpar{Training details:}
We adhere to the standard training protocols. Below are the references:

\begin{figure}[h!]
\begin{center}
\scalebox{0.9}{
\begin{tabular}{l|l} 
    \toprule
    Vision training code & \href{https://github.com/pytorch/vision/tree/main/references/classification}{PyTorch TorchVision references} \\
    ViT implementation & \href{https://github.com/pytorch/vision/tree/main/torchvision/models}{PyTorch TorchVision models} \\
    MLP-mixer implementations & \href{https://github.com/huggingface/pytorch-image-models/blob/main/timm/models/mlp_mixer.py}{huggingface/pytorch-image-models} \\
    Quantization & \href{https://github.com/TimDettmers/bitsandbytes}{TimDettmers/bitsandbytes} \\
    \bottomrule
\end{tabular}
}
\end{center}
\end{figure}

We replace the fused linear layers with standard linear layers to use LoRA. LoRA is applied across all linear layers. All experiments incorporate mixed-precision training. For nodes equipped with 4 GPU devices, we implement gradient checkpointing. We used gradient checkpointing for ViT-B and ViT-L.

\xpar{Hardware:}
Our experiments were conducted using various NVIDIA GPUs, including V100 and Titan RTX. 

\xpar{Architecture detail:}

\begin{table}[h!]
\begin{center}
\scalebox{0.85}{
\begin{tabular}{l|ccc} 
    \toprule
    Architecture     & ViT-S & ViT-B & ViT-L \\
    \midrule
    Patch-sizse      & 32    & 32    & 32  \\
    Attention blocks & 12    & 12    & 24 \\
    Attention heads  & 6     & 12    & 16 \\
    Hidden dim       & 6     & 768   & 1024 \\
    MLP dim          & 1536  & 3072  & 4096 \\
    \midrule
    Total parameters & 22.9M & 88.2M & 306.5M \\
\bottomrule
\end{tabular}
}
\caption{\small ViT architecture details.}
\label{table:vit-cfg}
\end{center}
\end{table}

\begin{table}[h!]
\begin{center}
\scalebox{0.85}{
\begin{tabular}{l|ccc} 
    \toprule
    Architecture     & Mixer-T & Mixer-S & Mixer-B \\
    \midrule
    Patch-sizs       &  32   & 32    & 32  \\
    Mixer blocks     &  6    & 8     & 12 \\
    Embed dim        & 384   & 512   & 768 \\
    MLP dim          & 1536  & 2048  & 3072 \\
    \midrule
    Total parameters &  8.8M & 19.1M & 60.3M  \\
\bottomrule
\end{tabular}
}
\caption{\small MLP-mixer architecture details}
\label{table:mlp-cfg}
\end{center}
\end{table}

\begin{table}[h!]
\begin{center}
\scalebox{0.85}{
\begin{tabular}{l|ccc} 
    \toprule
    Architecture     & NanoGPT & GPT2 \\
    \midrule
    Block-size       & 256     & 1024     \\
    Attention blocks & 6       & 12    \\
    Attention heads  & 6       & 12   \\
    Hidden dim       & 384     & 768   \\
    MLP dim          & 1152    & 2304  \\
    \midrule
    Total parameters & 10.7M   & 124.4M \\
\bottomrule
\end{tabular}
}
\caption{\small LLM GPT architecture details.}
\label{table:vit-cfg}
\end{center}
\end{table}

\newpage
\xpar{Dataset details:}
\label{app:datasets}

\begin{table}[h!]
\begin{center}
\scalebox{0.85}{
\begin{tabular}{l|ccccccc} 
    \toprule
    Dataset           & CIFAR10 & CIFAR100 & STL10 & CALTECH256 & SUN397 & ImageNet100 & ImageNet1K \\
    \midrule
    Original image-size  & $32 \times 32$ & $32 \times 32$ & $96 \times 96$ & Variable & Variable & Variable & Variable \\
    Training image-size & $224 \times 224$ & $224 \times 224$ & $224 \times 224$ & $224 \times 224$ & $224 \times 224$ & $224 \times 224$ & $224 \times 224$ \\
    Number of classes & $10$ & $100$ & $10$ & $257$ & $397$ & $100$ & $1,000$ \\
    Number of images  & $60,000$ & $60,000$ & $13,000$ & $30,607$ & $108,754$ & $130K$ & $>$1.2M \\
    \midrule
    Learning-rate $\eta_{\mathsf{default}}$ & $3 \cdot 10^{-4}$ & $3 \cdot 10^{-4}$ & $3 \cdot 10^{-4}$ & $3 \cdot 10^{-4}$ &  $3 \cdot 10^{-3}$ &  $3 \cdot 10^{-3}$ &  $3 \cdot 10^{-3}$\\
    Learning-rate  $\eta_{\mathsf{lte}}$ & $2 \cdot 10^{-5}$ & $2 \cdot 10^{-5}$ & $2 \cdot 10^{-5}$ & $2 \cdot 10^{-5}$ & $5 \cdot 10^{-6}$ & $3 \cdot 10^{-5}$ & $3 \cdot 10^{-5}$\\
    Batch-size & $1024$ & $1024$ & $1024$ & $1024$ & $1024$ & $4096$ & $8192$\\
\bottomrule
\end{tabular}
}
\caption{\small Specifications for vision datasets. Most dataset images indicated as ``variable size'' are larger than the training image size. We provide training configuration used for ViT. For MLP-Mixer on ImageNet100, we use a learning rate of $0.001$ for full-model pre-training and $1 \cdot 10^{-4}$ for LTE, both with a batch size of 4096.} 
\label{table:dataset-specs}
\end{center}
\end{table}

\vspace{-0.2in}

\begin{table}[h!]
\begin{center}
\scalebox{0.85}{
\begin{tabular}{l|cc} 
    \toprule
    Dataset                 & Shakespeare & TinyStories \\
    \midrule
    Total number of tokens  & $1.0M$ & $474.0M$ \\
    Tokenizer size          & $65$   & $50304$  \\
    \midrule
    Learning-rate $\eta_{\mathsf{default}}$ & $1 \cdot 10^{-3}$ & $6 \cdot 10^{-4}$\\
    Learning-rate  $\eta_{\mathsf{lte}}$ & $2 \cdot 10^{-6}$ & $5 \cdot 10^{-5}$ \\
    Batch-size & $512$ & $512$ \\
    Block-size & $256$ & $1024$ \\
    \bottomrule
\end{tabular}
}
\caption{\small Specifications for LLM datasets and hyper-parameters used for miniGPT on Shakespeare and GPT2 on Tinystories.}
\label{table:llm-training-specs}
\end{center}
\end{table}

\xpar{LTE Optimization Details:}
We use \(\alpha=4096\sim8192\), which is \(s=128\sim256\) when \(r=32\) and \(s=64\sim128\) when \(r=64\). A good rule of thumb for the learning rate is to set it at approximately \(0.1 \sim 0.05\times\) the standard model training learning rate. We use the same learning rate scheduler as the standard pre-training, which is cosine learning-rate decay with linear warmup. 

\xpar{LTE Batching Detail:}
We use a fixed cumulative batch size for LTE. This means that given a batch size \(B\) with \(N\) LoRA heads, each head receives a batch size of \([B/N]\). When counting the training iterations, we count \(B\) and not \([B/N]\). Counting using \([B/N]\) would significantly inflate our numbers, but in the federated learning community, it is referred to as ``synchronization steps''. LTE training epochs were set to \(4\times\) the cumulative batch size, and we exited early when we matched the performance of full-model training. For smaller datasets, our method seemed to consistently outperform the baseline, likely due to the regularization properties of rank and over-parameterization.

\xpar{LTE Implementation Details:}
We implemented LTE using Parameter Server, model-parallelism, and DDP. Parameter server is theoretically more beneficial for our method, we conducted most of our development using DMP and DDP as it does not require rewriting communication logic. We utilize `\texttt{torch.vmap}' and simulate multiple devices on the same GPU, and share computation of the main parameters to reduce run-time.

\xpar{LTE for Convolution, Affine, and Embedding Layers:}
Specific choices for all these layers did not impact the final performance significantly, but we detail the choices we made below.

For convolution layers, we use the over-parameterization trick in~\citep{huh2023simplicitybias}, which uses \(1\times1\) for the second layer. Since convolution layers are typically used at most once in the models we tested, we did not explore beyond this parameterization. However, other potential choices exist for low-rank parameterization of convolution layers, such as channel-wise convolution and separable convolutions.

There is no notion of low-rank decomposition for affine parameters used in normalization layers. We tried various strategies to train and communicate these parameters. For affine parameters, we tried: (1) LoRA-style vector-vector parameterization \(\textbf{a}, \textbf{b}\), (2) LoRA-style vector-scalar parameterization \(\textbf{A} = \textbf{a}, b\), (3) DDP-style averaging, and (4) removing affine parameters. Removing affine parameters improved both baseline and LTE for classification.

Lastly, we found that using the standard averaging technique or allowing only one model to train the embedding layer worked best for the embedding layer. We chose to use standard averaging at the same iteration as the rest of the LoRA layers.

\subsection{Getting exact equivalence}
\label{app:exact}
The exact equivalence condition for LTE and MHLoRA is achieved when the LoRA parameters are not reset. Empirically, this does lead to slightly better model performance but requires a more involved calculation. We posit that the slight improvement comes from the fact that we do not have to re-learn the LoRA parameters and ensure the optimizer state is consistent with respect to its parameters. We provide the exact equivalence below.

Denote $t$ as the synchronization steps, $\tau$ as the local optimization step:

\begin{flalign*}
\underset{\bB_0, \bA_0, \dots \bB_N, \bA_N}{\text{optimize}} \quad \bW \bx + \frac{s}{N} \sum_{i=1}^N \bB_i^{(t, \tau)} \bA_i^{(t, \tau)}\bx
\end{flalign*}

To match the gradient dynamics exactly, one needs to merge all LoRA and subtract the contribution of its weight after the merge. Denote $\bV$ as the previous LoRA parameter contribution and is set to $\mathbf{0}$. Then, the LTE optimization with the correction term is:

\begin{tcolorbox}[colback=red!1!white,colframe=white!75!black,boxrule=2pt,arc=3.4pt,boxsep=-1mm,left=0pt,right=0pt,top=0pt,bottom=10pt]
\begin{flalign*}
& (\text{For each}\; \bB_i, \bA_i \; \text{optimize for} \; \tau \; \text{steps})\; \\
& \qquad\qquad\qquad   \bW^{(t)} \bx - s\bV_i^{(t)} \bx + s\bB_i^{(t, 0)} \bA_i^{(t, 0)} \bx  \\
& (\text{Merge all} \; \bB_i, \bA_i) \\
& \qquad \bW^{(t+1)} = \bW^{(t)} + \frac{s}{N} \sum_{i=1}^N  \left( \bB_i^{(t, \tau)} \bA_i^{(t, \tau)} - \bV_i^{(t)}\right) \\
& (\text{Update } \bV_i ) \\
& \qquad\qquad\qquad\qquad\qquad\qquad   \bV_i^{(t+1)} = \bB_i^{(t, \tau)} \bA_i^{(t, \tau)} \\
& (\text{Use same parameters} ) \\
& \qquad\qquad\qquad\qquad   \bB_i^{(t+1, 0)}, \bA_i^{(t+1, 0)}  = \bB_i^{(t, \tau)}, \bA_i^{(t, \tau)} 
\end{flalign*}
\end{tcolorbox}

Where the equivalence to multi-head LoRA holds when $\tau=1$. Note that the subtracting of the previous contribution can be absorbed into the weight $(\bW^{(t)} - s\bV_i^{(t)})\bx$.

\newpage
\subsection{Merge with least-squares}
\label{app:merge-least-squares}
\begin{figure}[h]
    \centering
    \includegraphics[width=0.495\linewidth]{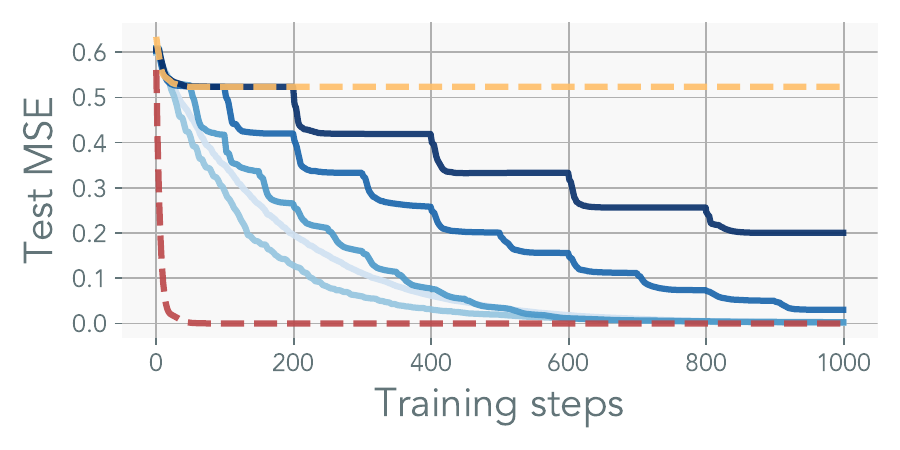}
    \hfill
    \includegraphics[width=0.495\linewidth]{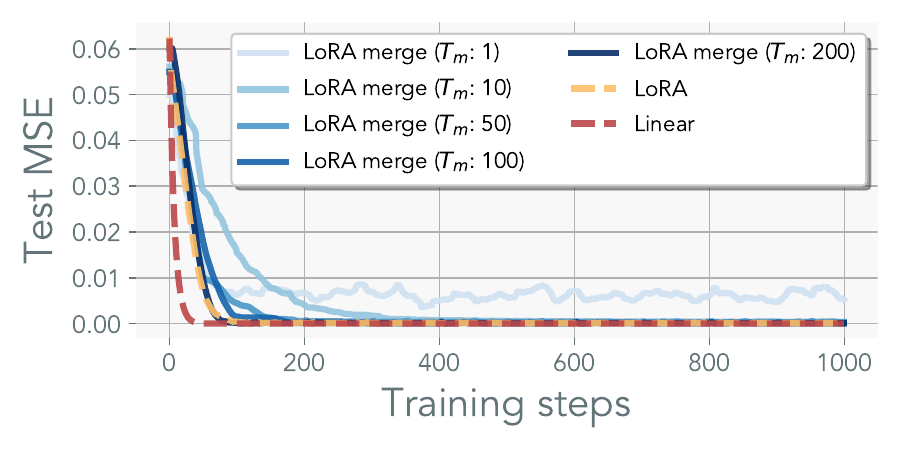}
    \caption{\small\textbf{Least-squares with LoRA}: Linear models parameterized with LoRA with varying target rank. Least-squares with $\mathbb{R}^{32\times32}$, with target rank of $32$~(\textbf{right}) and $8$~(\textbf{left}). LoRA is parameterized with rank $r=4$. With merging, the model can recover the solution, with convergence scaling with merge frequencies.}
    \label{fig:merge_ls}
\end{figure}

We train a linear network, parameterized with LoRA, on least-squares regression. Here, we artificially constructed the problem to control for the underlying rank of the solution $\bW^*$. We then constructed a dataset by randomly generating $\bY = \bX (\bW^*)^{\mathsf{T}}$. Where for each element $\bx \in \bX$ is drawn from a normal distribution $\bx \sim \mathcal{N}(\mathbf{0}, \mathbf{I})$.

Figure \ref{fig:merge_ls} visualizes the model's ability to recover the underlying ground-truth solution across various merge iterations $T$. Here, the optimal solution $\bW^*$ is set to be full rank where $m=n=32$. 
We employ a naive re-initialization strategy of initializing $\bA$ with a uniform distribution scaled by the fan-out.

Without merging the LoRA parameters, the model's performance rapidly plateaus. In contrast, models trained with merges can eventually recover the full-rank solution, with the recovery rate of the solution scaling with the merging frequency.

\newpage
\subsection{Ablation}
\label{app:ablation}
\begin{figure}[h]
    \centering
    \subfigure[Initialization]{\label{subfig:init_ablation_losses}\includegraphics[width=0.32\linewidth]{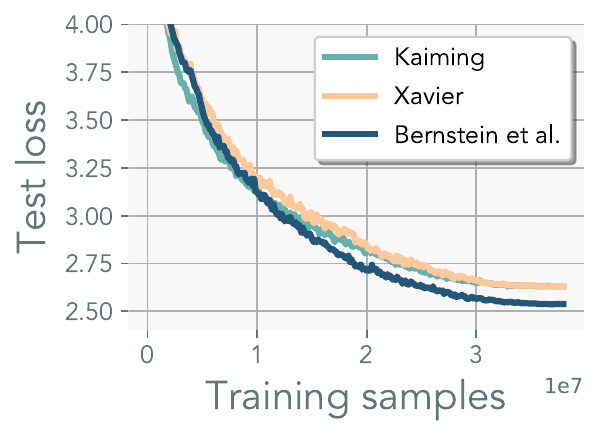}}
    \subfigure[LoRA $\bA$ reset]{\label{subfig:A_reset_ablation}\includegraphics[width=0.32\linewidth]{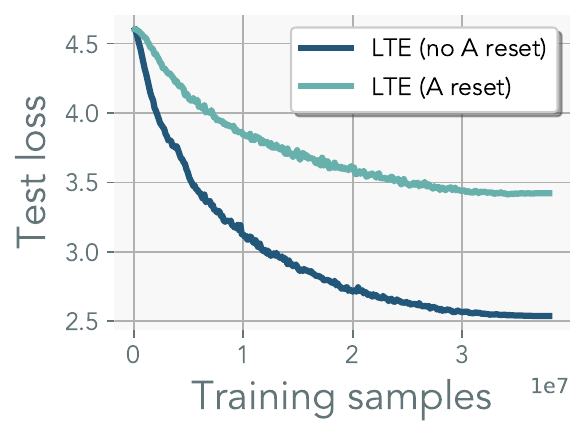}}
    \subfigure[Optimizer reset]{\label{subfig:opt_reset_ablation}\includegraphics[width=0.32\linewidth]{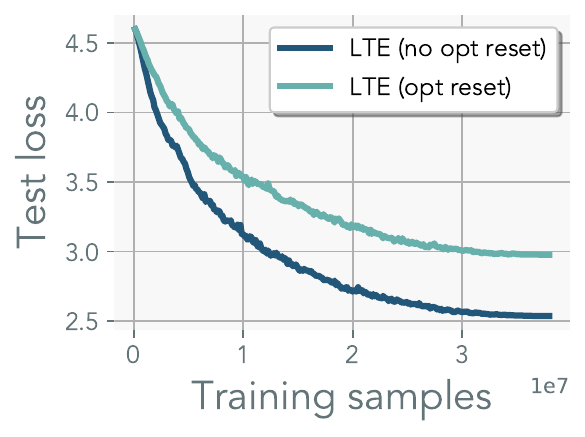}}
    \caption{\textbf{Ablation}: These models were trained using ViT-S with $8$ heads with rank $r=16$. (\textbf{Left}) different initialization scheme. (\textbf{Middle}) resetting $\bA$. (\textbf{Right}) resetting optimizer states for both $\bA$ and $\bB$.}
    \label{fig:ablation}
\end{figure}

We conducted an ablation study focusing on initialization, resetting of LoRA \( \mathbf{A} \), and resetting the optimizer for the LoRA parameters. All ablation studies presented here were conducted with an LTE with rank $r=16$, $8$ heads using ViT-S on ImageNet100.

Kaiming initialization serves as the default scheme for LoRA. The conventional Kaiming initialization is tailored for square matrices and depends solely on the input dimension. Given that LoRA parameters often manifest as wide matrices, we experimented with various initialization schemes. Xavier initialization~(\cite{glorot2010understanding}) preserves the variance relationship of a linear layer for rectangular matrices. Bernstein et al.~(\cite{bernstein2023agd}) employ semi-orthogonal initialization to maintain the spectral norm of the input. As depicted in Figure~\ref{subfig:init_ablation_losses}, the method by Bernstein et al. proved most effective. It should be noted that none of these initialization methods assume residual connection or zero-ed out LoRA parameters. Hence, further tuning of the gain parameters might be needed.

When resetting the LoRA parameters, we investigated the impact of re-initializing the matrix $\mathbf{A}$ as well as its optimizer. As shown in~\fig{subfig:A_reset_ablation}, we found that resetting matrix $ \mathbf{A}$ adversely affects model performance, possibly due to the necessity of re-learning the representation at each iteration and discarding the momentum states. In~\fig{subfig:opt_reset_ablation}, we also experimented with retaining the LoRA parameters while resetting the optimizer state for those parameters. Similarly, we found that resetting the optimizer state diminishes performance.

\newpage
\subsection{Grassman distance of LoRA heads}
\label{app:grass}
Motivated by~\citep{hu2021lora}, we measure the Grassman distance of the LoRA heads to measure sub-space similarity between the optimized sub-spaces. Grassman distance measures the distance of $k$-dimensional subspace in a $n$-dimensional space. The distance is defined as:

\begin{tcolorbox}[colback=white,colframe=white!75!black,boxrule=2pt,arc=2pt,boxsep=-1mm,left=8pt,right=8pt,top=10pt,bottom=10pt]
\xpar{Grassman distance} Given subspaces \( U \) and \( V \), and given the singular values \( U^T V = X \Sigma Y^T \), where \( \Sigma \) is a diagonal matrix with singular values \( \sigma_i \). The principal angles \( \theta_i \) between \( U \) and \( V \) are given by \( \theta_i = \cos^{-1}(\sigma_i) \). Then the Grassmann distance \( d_{\mathsf{Grassman}}(U, V) \) is then defined as:

\[ d_{\mathsf{Grassman}}(U, V, k) = \left( \sum_{i=1}^{k} \theta_i^2 \right)^{\frac{1}{2}} \]

Where \( k \) is the number of principal angles, the dimension of the smaller subspace. 
\end{tcolorbox}

For LoRA, the number of principal components is spanned by the LoRA rank $r$; therefore, we set $k=r$. When the LoRA parameters span the same sub-space, they have a Grassman distance of $0$. The pairwise Grassman distance is measured by:
\begin{align}
\frac{1}{2N}  \hspace{-0.05in}  \sum_{(i,j) \in [1, N]  \land i \neq j} \hspace{-0.2in} d_{\mathsf{Grassman}}(f_{\perp}(\bB_i\bA_i), f_{\perp}(\bB_j\bA_j), r) 
\end{align}

Where $f_{\perp}(\cdot)$ computes the orthogonal basis by computing the input matrix's left singular vectors.

\newpage
\subsection{Training curve on ImageNet1k}
\label{app:imagenet1k_test_curve}
\begin{figure}[h]
    \centering
    \includegraphics[width=1.0\linewidth]{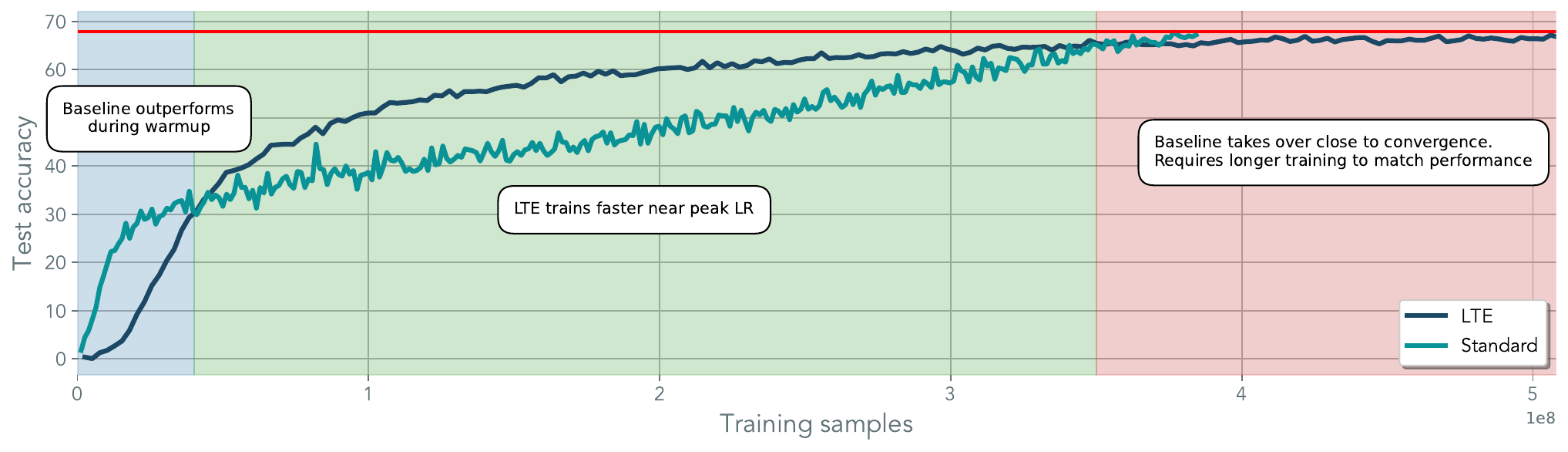}
    \caption{\small\textbf{ImageNet1k test curve}}
    \label{fig:imagenet1k_test_curve}
\end{figure}

We plot the test curves in~\fig{fig:imagenet1k_test_curve} for both standard and LTE training on ImageNet-1K. Both models were trained using a cosine learning rate scheduler. We found that training LTE for $300$ epochs performed roughly $5\%$ worse. Hence, we repeated the experiment by setting the training epoch to $600$. The final performance was matched at around $420$ epochs. For LTE, we doubled the batch size to $8192$ (see~\app{app:batch_size_imagenet1k}). We found that LTE benefits from larger cumulative batch size and does not require as intensive of data augmentation as standard training, likely due to implicit regularization induced by low-rank parameterization and stale updates. The baseline did not improve when using a bigger batch size. In simpler tasks, using bigger-batches did not perform better.

Early in the training phase, the baseline outperformed LTE, but this trend was quickly reversed after the first initial epochs. LTE approached its final performance quite rapidly. However, LTE fell short compared to the standard training duration by 300 epochs, and an additional $120$ epochs were required to reach the same test accuracy. Unlike LTE, we found that standard training benefited significantly more from the learning-rate scheduler, possibly hinting that the gradient noise and stale parameter estimates overpower the signal-to-noise ratio. We observed this trend across all ViT sizes. 

A few ways to mitigate the slow convergence include synchronizing the mini-batches or the LoRA parameters as the model is trained; we leave such investigation for future works. The baseline was trained with a weight-decay of $0.01$, resulting in slow convergence but better final performance -- standard optimization does exhibit faster convergence if the weight-decay is removed. Training the baseline for longer, $600$ epochs, did not achieve better performance. 

\newpage
\subsection{Effect of batch-size on ImageNet1k}
\label{app:batch_size_imagenet1k}
\begin{figure}[h]
    \centering
    \includegraphics[width=0.495\linewidth]{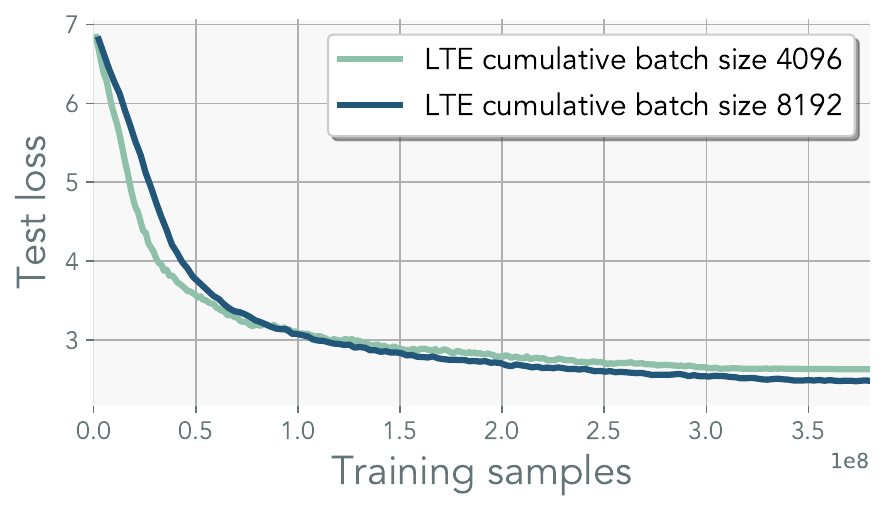}
    \hfill
    \includegraphics[width=0.495\linewidth]{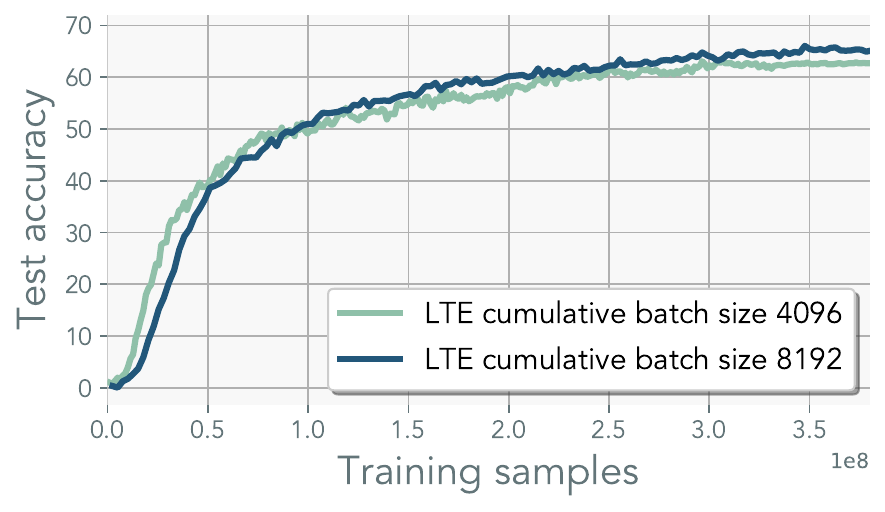}
    \caption{\small\textbf{ImageNet1k with doubled accumulative batch-size}}
    \label{fig:imagenet_batch_size}
\end{figure}

Utilizing larger batch sizes is beneficial for maximizing FLOP efficiency. However, when evaluated in terms of samples seen, larger batch sizes are known to underperform~\cite{masters2018revisiting}. Given that LTE introduces more noise, we hypothesized that a reduced learning rate could be adversely affected by both gradient noise and estimate noise from using merges. In Figure~\ref{fig:imagenet_batch_size}, we experimented with increasing the batch size and observed a moderate improvement of $3\%$ with bigger batches.

\newpage
\section{Rank of the model in pre-training}
\label{app:vit-rank-dynamics}
\begin{figure}[h]
    \centering
    \includegraphics[width=1.0\linewidth]{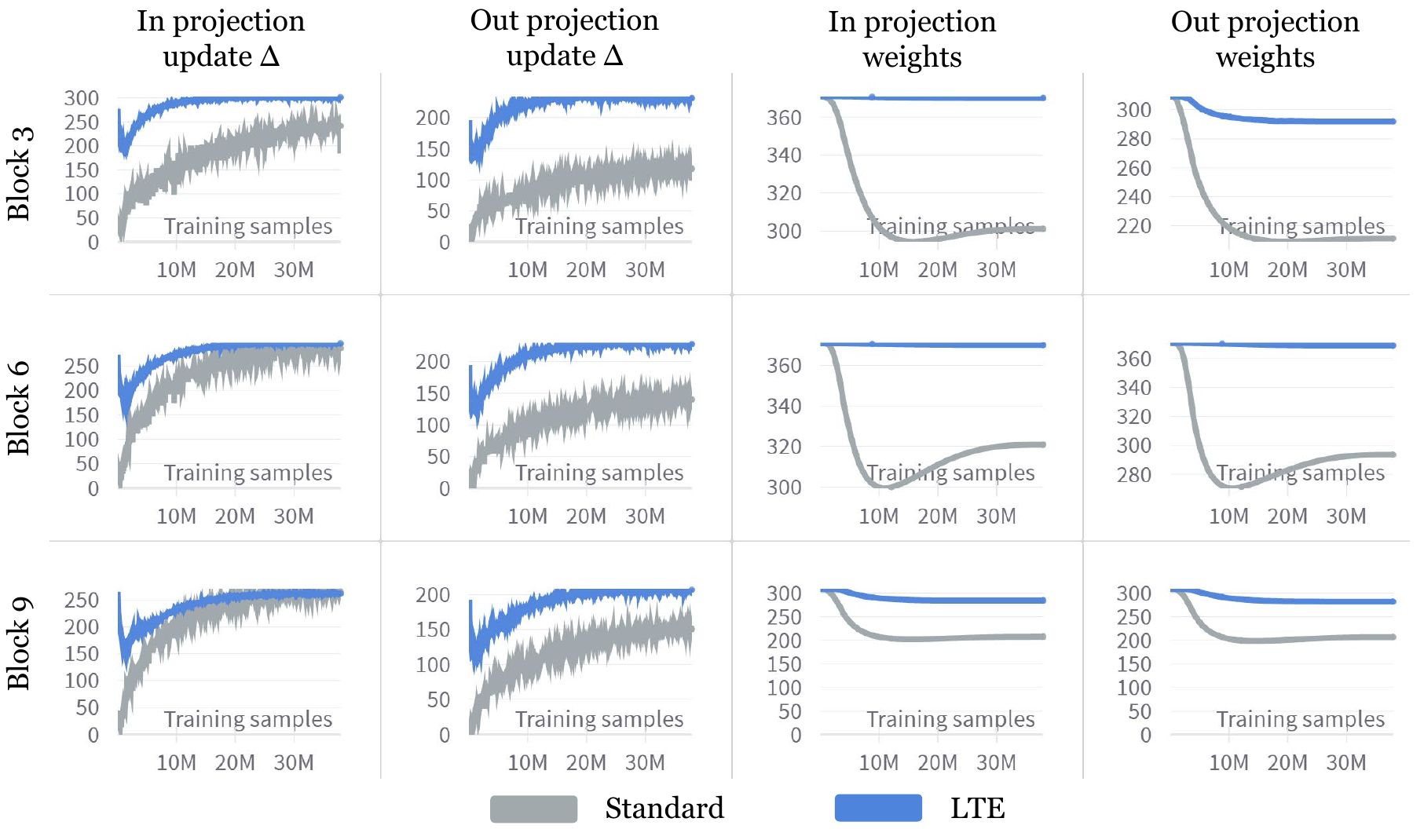}
    \caption{\textbf{Rank dynamics of ViT for standard training and LTE}. Rank is measured using effective rank. We track the rank of the weights and update to the main weight throughout training.}
    \label{fig:stable_rank}
\end{figure}

We measure the effective rank~\citep{roy2007effective} of standard training and LTE throughout training.

\begin{tcolorbox}[colback=white,colframe=white!75!black,boxrule=2pt,arc=2pt,boxsep=-1mm,left=8pt,right=8pt,top=10pt,bottom=10pt]
\xpar{(Definition) Effective rank (spectral rank)} 
For any matrix $A \in \mathbb{R}^{m \times n}$, the effective rank $\rho$ is defined as the Shannon entropy of the normalized singular values:
\[ \rho(A) = \text{exp}\left( -\sum_{i=1}^{\min(n, m)} \bar \sigma_i \log( \bar \sigma_i) \right), \]
where $ \bar \sigma_i =  \sigma_i/\sum_{j} \sigma_j $ are normalized singular values, such that $\sum_i \bar \sigma_i = 1$. 
\end{tcolorbox}

The rank of the updates for standard training is the gradients, and LTE is $\nabla_\bW \cL $ and $\frac{s}{N}\sum_n \bB_n\bA_n$ respectively.
In the context of standard training, the rank of the weights exhibits only a minor decrease throughout the optimization process. Conversely, the rank of the gradient monotonically increases following the initial epochs. This observation provides empirical evidence that approximating the updates with a single LoRA head is not feasible. Despite its markedly different dynamics, LTE can represent full-rank updates throughout the training period. This may also be useful for designing how many LoRA heads to use with LTE, where the number of LoRA heads can start with one and slowly annealed to match the maximum rank of the weights.

\newpage
\subsection{Is scaling \texorpdfstring{$s$}{s} the same as scaling the learning rate?}
\label{app:s_vs_lr}
There is a misconception that the scalar $s$ only acts as a way to tune the learning rate in these updates. 
Focusing on the update for $\bB$ (same analysis holds for $\bA$), we can write out the gradient as:

\begin{align}
g(\bB) = \frac{\partial \cL}{\partial \bB} = s\frac{\partial \cL}{\partial \bz_{out}} (\bA \bz_{in})^T = s \bar{g}_t
\end{align}

If we were using stochastic gradient descent, we would expect $s$ to behave like a linear scaling on the learning rate:
\begin{align}
\Delta(\bB) = -\eta s \bar g
\end{align}

Where we denoted $\bar g$ as the component of the gradient with $s$ factored out. We now show that $s$ does not linearly scale the learning rate for Adam (this analysis can be extended to scale-invariant optimizers). Adam is a function of the first-order momentum $m_t$ and second-order momentum $v_t$. One can factor out $s$ from the momentum term: $m_t = s \hat{m}_t = s \beta_1 \hat{m}_{t-1} + (1 -\beta_1) \hat{g}_t $  and $v_t = s^2 \hat{v}_t = s^2 \beta_1 \hat{v}_{t-1} + (1 -\beta_1) \hat{g}_t^2 $. 
Incorporating the gradient into the update rule, we see that the adaptive update does not depend linearly on $s$:
\begin{align}
\Delta(\bB) 
= -\eta \frac{m_{t}}{\sqrt{ v_{t} + \epsilon }}
= -\eta \frac{ {\color{red} s} \hat{m}_{t}}{\sqrt{ {\color{red} s^2} \hat{v}_{t} + \epsilon }} = -\eta \frac{\hat{m}_{t}}{\sqrt{\hat{v}_{t} + \epsilon }}
\end{align}

However, $\hat{g}_t$ is not invariant to $s$. Therefore, while $s$ is not the same as the learning rate, it will impact the downward gradient $\frac{\partial \cL(\dots, {\color{red} s} )}{\partial \bz_{out}}$. We will discuss this in the next subsection. It is worth noting that $s$ quadratically impacts the attention map and may require a separate scaling factor as opposed between MLP layers and attention layers. Consider a batched input $\bX$ with the output of a linear as $\hat\bX = \bW \bX + {\color{red} s} \bB \bA \bX = \bW \bX + s \mathbf{D}$. Then, the un-normalized attention map is dominated by the LoRA parameters:
\begin{align}
\left( \bW_Q \hat X \right) \left( \bW_K \hat X \right)^T &= \bW_Q \left( \bW \bX + s \mathbf{D} \right) \left(  \bX^T \bW^T + s \mathbf{D}^T \right)  \bW_K^T \\
& = \dots 
+ {\color{red} s^2} \bW_Q \mathbf{D} \mathbf{D}^T  \bW_K^T
\end{align}

Unlike learning rate, scaling $s$ affects both the forward and backward dynamics. Large $s$ emphasizes the contribution of the LoRA parameters, which may explain why we have observed better performance when using larger $s$ for pre-training. It is possible that using a scheduler for $s$ could further speed up training or even better understand how to fuse $s$ into the optimizer or $\bA$; we leave this for future work. Next, we dive into the effect of $s$ on $\bar{g}_t$.

\newpage
\subsection{The effective update rule for LoRA is different from standard update}
\label{app:lora_effective_update}
\begin{tcolorbox}[colback=white,colframe=white!75!black,boxrule=2pt,arc=2pt,boxsep=-1mm,left=8pt,right=8pt,top=10pt,bottom=10pt]
\xpar{Effective update of LoRA.} 
Let $\bW$ be the original weight of the model, and denote $g(\bW)=g$ as the gradient of the parameter. Let $\hat \bW = \bW + s \bB \bA$ be the effective weight of the LoRA parameterization, and $g(\hat\bW) = \hat g$ be its corresponding effective gradient. Then, the LoRA parameterization is related to the gradient of the standard parameterization by 
\begin{align}
\label{eqn:lora-update}
\hat g = s \left( \bB \bB^T g -  g \bA^T \bA \right)  - s^2 \eta  \left( g \left( \bB\bA \right)^T g \right)
\end{align}
\end{tcolorbox}

When $s$ is small, we can safely discard the second term as it will scale quadratically with learning rate $\eta \hat g$. 
However, when $s$ is large, the contribution of the second term becomes non-negligible. This term can be interpreted as the alignment of the LoRA parameters, and taking a step in this direction encourages $\bB$ and $\bA$ to be spectrally aligned. The increased contribution of the LoRA parameters and the alignment induced by larger $s$ may explain our observation that higher $s$ leads to better performance. It's important to note that with a learning rate scheduler, the contribution of the second term would decay to zero. 

\subsection{Derivation}
\label{app:effective-update}
Over-parameterization or linear re-parameterization, in general, has a non-trivial effect on the optimization dynamics. 
Here, we analyze the update of the effective weight to point out a rather surprising interaction between $s$ and $\eta$. Consider a standard update rule for SGD for $\bz_{in} \in \mathbb{R}^{n \times 1}$, and $\bz_{out} \in \mathbb{R}^{m \times 1}$ and $\bW \in \mathbb{R}^{m\times n}$:

\begin{align}
g(\bW) &= \frac{\partial \cL}{\partial \bW} 
= \frac{\partial \cL }{ \partial \bz_{out}}  \frac{\partial \bz_{out}}{\partial \bW}  
= \frac{\partial\cL}{\partial\bz_{out}}  \bz_{in}^T
\end{align}

We will denote $g(\bW) = g$ from now on for clarity. For standard LoRA with parameters $\hat \bW = \bW + s \bB \bA$, where $\bB \in \mathbb{R}^{m\times r}$, $\bA \in \mathbb{R}^{r\times n}$, the update rule on the effective weight is: 

\begin{align}
\hat \bW &\leftarrow \bW + s \left( \bB - \eta \frac{\partial \cL}{\partial \bB} \right) \left(\bA - \eta \frac{\partial \cL}{\partial \bA} \right) \\
&= \bW + s \bB\bA - s\eta \left( \left( \bB \frac{\partial \cL}{\partial \bA} - \bA \frac{\partial \cL}{\partial \bB} \right) +  \eta \frac{\partial \cL}{\partial \bB}\frac{\partial \cL}{\partial \bA } \right)  
\end{align}

We denote $g(\hat \bW)$ as $\hat g$. With the resulting effective update being:

\begin{align}
\hat g =  \left( \bB \frac{\partial \cL}{\partial \bA} - \frac{\partial \cL}{\partial \bB} \bA \right) - \eta \frac{\partial \cL}{\partial \bB}\frac{\partial \cL}{\partial \bA} 
\end{align}

Computing the derivative for each variable introduces the dependency on $s$.
\begin{align}
\frac{\partial \cL}{\partial \bB} &= \frac{\partial \cL}{\partial \bz_{out}} \frac{\partial \bz_{out}}{\partial \bz_{res}} \frac{\partial \bz_{res}}{\partial \bB}
= s \frac{\partial \cL}{\partial \bz_{out}} (\bA \bz_{in})^T \\
\frac{\partial \cL}{\partial \bA} &= \frac{\partial \cL}{\partial \bz_{out}} \frac{\partial \bz_{out}}{\partial \bz_{res}} \frac{\partial \bz_{res}}{\partial \bA}
= s \bB^T \frac{\partial \cL}{\partial \bz_{out}} \bz_{in}^T 
\end{align}

Plugging it back in, we have\begin{align}
\hat g &=  \left( 
\bB 
\left( s \bB^T \frac{\partial \cL}{\partial \bz_{out}} \bz_{in}^T \right)
- 
\left( s \frac{\partial \cL}{\partial \bz_{out}} (\bA \bz_{in})^T \right)
\right) 
\bA
- \eta 
\left( s \frac{\partial \cL}{\partial \bz_{out}} (\bA \bz_{in})^T \right)
\left( s \bB^T \frac{\partial \cL}{\partial \bz_{out}} \bz_{in}^T \right)\\
&= s \left( 
 \frac{\partial \cL}{\partial \bz_{out}} \bz_{in}^T \bA^T \bA 
 -
 \bB \bB^T \frac{\partial \cL}{\partial \bz_{out}} \bz_{in}^T
 \right) 
- s^2 \eta 
\left( \frac{\partial \cL}{\partial \bz_{out}} \bz_{in}^T \bA^T  \bB^T \frac{\partial \cL}{\partial \bz_{out}} \bz_{in}^T \right) \\
&= s \left( 
\bB \bB^T g
- 
g \bA^T \bA \right) 
- s^2 \eta 
\left( g \bA^T  \bB^T g \right)
\end{align}

When $s$ is small, both terms exist. When $s$ is large, the second term dominates. Since the last term is quadratic with $g$, one can safely ignore the second term when the learning rate is sufficiently small. Similarly, when using a learning rate scheduler, the contribution of the second term would decay to zero. 
The second term can be interpreted as an alignment loss where the gradient moves in the direction that aligns LoRA parameters.

\newpage
\section{Method illustrations}
In our illustrations, we detail the distinctions between our method and other common strategies. Distributed Data-Parallel (DDP) synchronizes the model at every iteration, with only the gradients being communicated between devices. This necessitates model synchronization across devices every iteration. Therefore, if there's a significant delay in synchronization due to slow interconnect speeds or large model sizes, synchronization becomes a bottleneck. One way to mitigate this is through local optimization, often referred to as local steps or local SGD in federated learning. Here, instead of communicating gradients, model weights are shared. Local steps are known to converge on expectation, but they still require communicating the full model, which is loaded in half or full precision, which will quickly become infeasible in $1$B+ size models.

Our proposed method addresses both communication and memory issues by utilizing LoRA. Each device loads a unique set of LoRA parameters, and these parameters are updated locally. As discussed in our work, this enabled efficient exploration of full-rank updates. We communicate only the LoRA parameters, which can be set to be the order of magnitude smaller than the original model's size. Our approach balances single-contiguous memory use with the ability to utilize more devices. The aim is to enable the training of large models using low-memory devices.

\begin{figure}[h]
    \centering
    \includegraphics[width=1.0\linewidth]{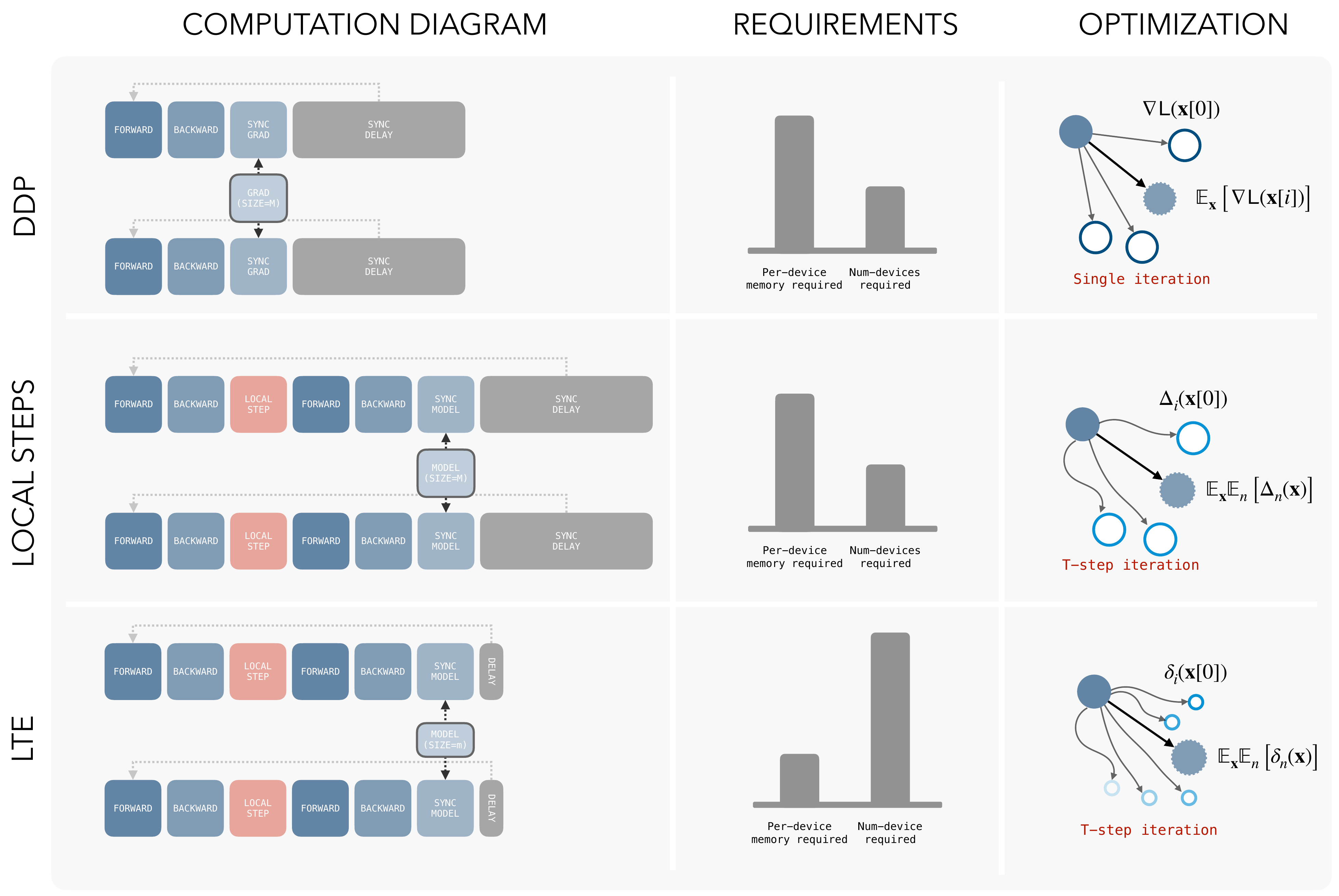}
    \caption{\textbf{Method illustration}: Comparisons between distributed learning methods to our method.
    }
    \label{fig:illustration}
\end{figure}

\newpage
\section{Additional results}
\label{app:more-results}

In all of our experiments, we present two distinct curves for analysis: one that represents the total training data observed across all devices and another that shows the training samples or tokens seen per device. We use LTE with $T=1$ for all experiments below, which is equivalent to the Multihead-LoRA optimization.

\subsection{Vision Image Classification}

We conduct additional experiments in image classification, covering datasets like CIFAR10~\citep{krizhevsky2009learning}, CIFAR100~\citep{krizhevsky2009learning}, STL10~\citep{coates2011analysis}, Caltech256~\citep{griffin2007caltech}, and SUN397~\citep{xiao2016sun}. For these tests, we re-tuned all baseline learning rates. Detailed information about these datasets is available in~\app{app:train-details}. For LTE, we use ranks of $r=64$ and $N=32$ heads.

\begin{figure}[htbp]
  \centering
\subfigure[Vision task test accuracy vs total training samples]{\label{fig:first}\includegraphics[width=\linewidth]{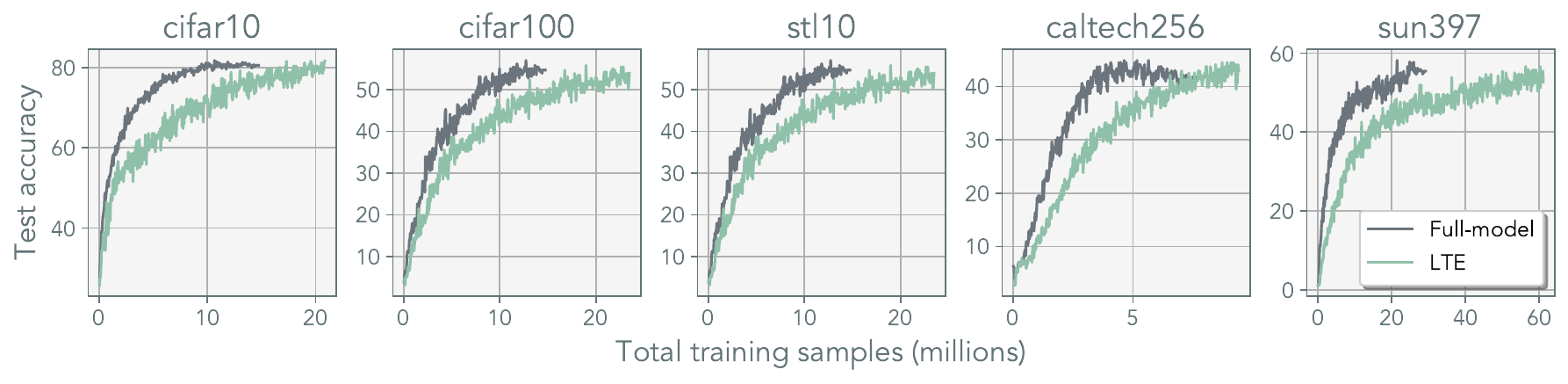}}\\
\subfigure[Vision task test accuracy vs training samples per device]{\label{fig:first}\includegraphics[width=\linewidth]{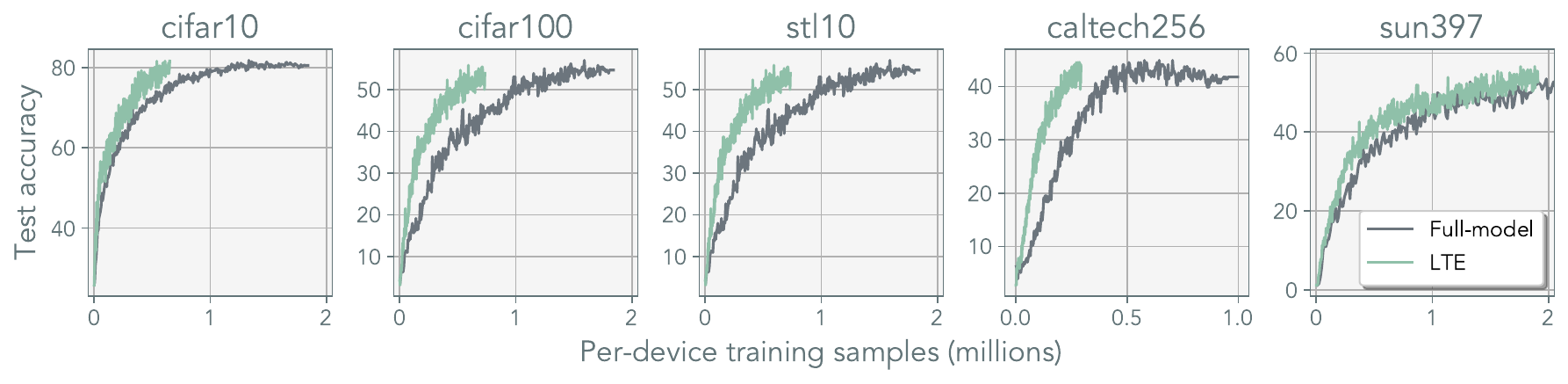}}
  \caption{\small \textbf{Additional results on various image classification datasets using ViT-S}}
  \label{fig:both_figures}
\end{figure}

\newpage
\subsection{Scaling up ViT model size}

We train larger variants of the Vision Transformer (ViT) model. Details on these architectures are provided in~\app{app:train-details}. Across all sizes, our results remained consistent, and for ViT-L, we used a rank of $r=128$.

\begin{figure}[htbp]
  \centering
\subfigure[ViT ImageNet100 test accuracy vs total training samples]{\label{fig:first}\includegraphics[width=\linewidth]{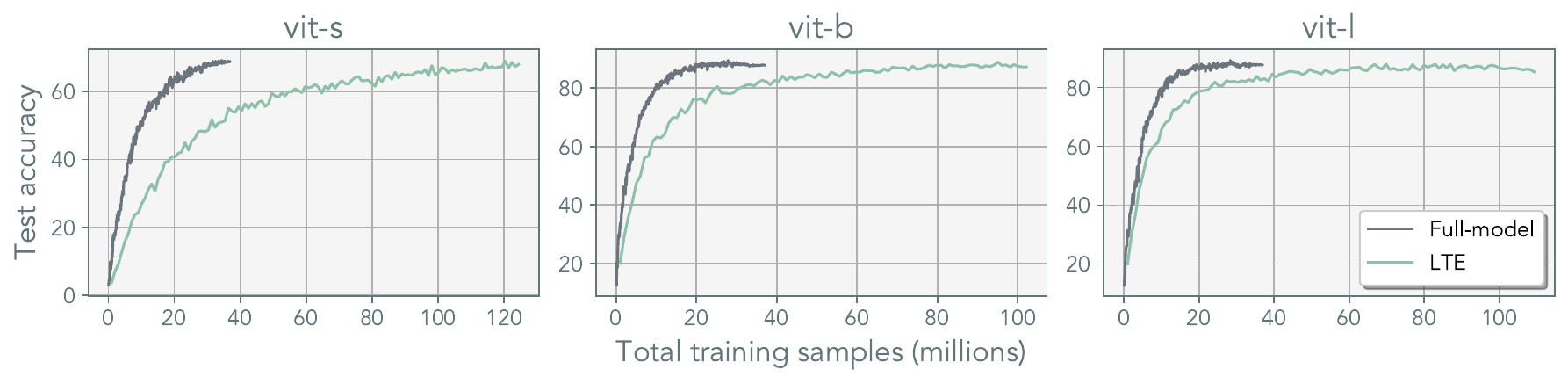}}\\
\subfigure[ViT ImageNet100 test accuracy vs training samples per device]{\label{fig:first}\includegraphics[width=\linewidth]{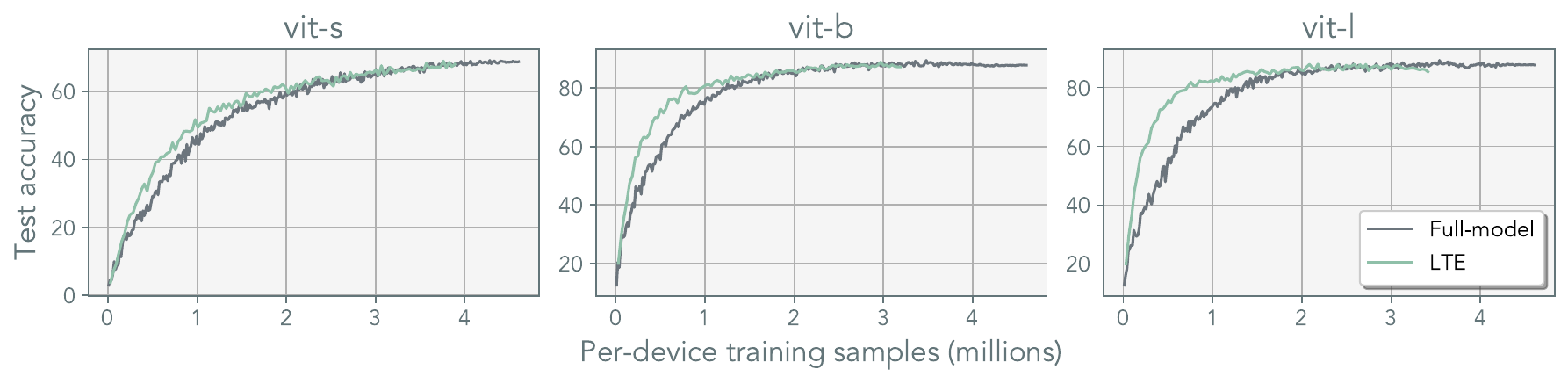}}
  \caption{\small \textbf{ImageNet100 classification on varying ViT scale}}
  \label{fig:both_figures}
\end{figure}

\newpage
\subsection{LTE on MLP-Mixer}
To evaluate the generalizability of our method to non-Transformer-based architectures, we train MLP-Mixer~\citep{tolstikhin2021mlp} using LTE. The specific details of the architecture used in datasets are listed in~\app{app:train-details}. Our findings are consistent results across different scales of the MLP-Mixer. For Mixer-B, we use a rank of $r=128$.

\begin{figure}[htbp]
  \centering
\subfigure[MLP-mixer ImageNet100 test accuracy vs total training samples]{\label{fig:first}\includegraphics[width=\linewidth]{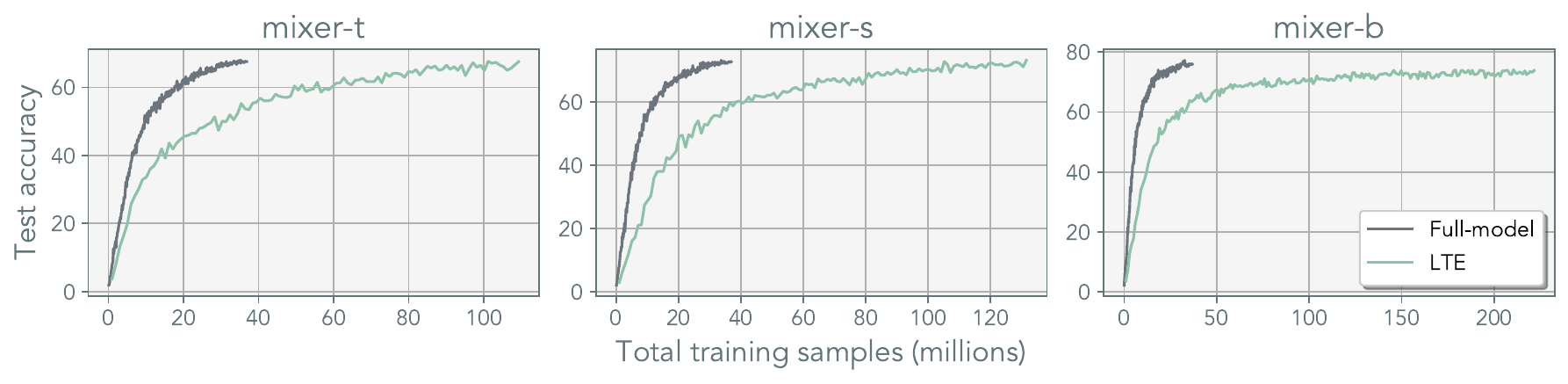}}\\
\subfigure[MLP-mixer ImageNet100 test accuracy vs training samples per device]{\label{fig:first}\includegraphics[width=\linewidth]{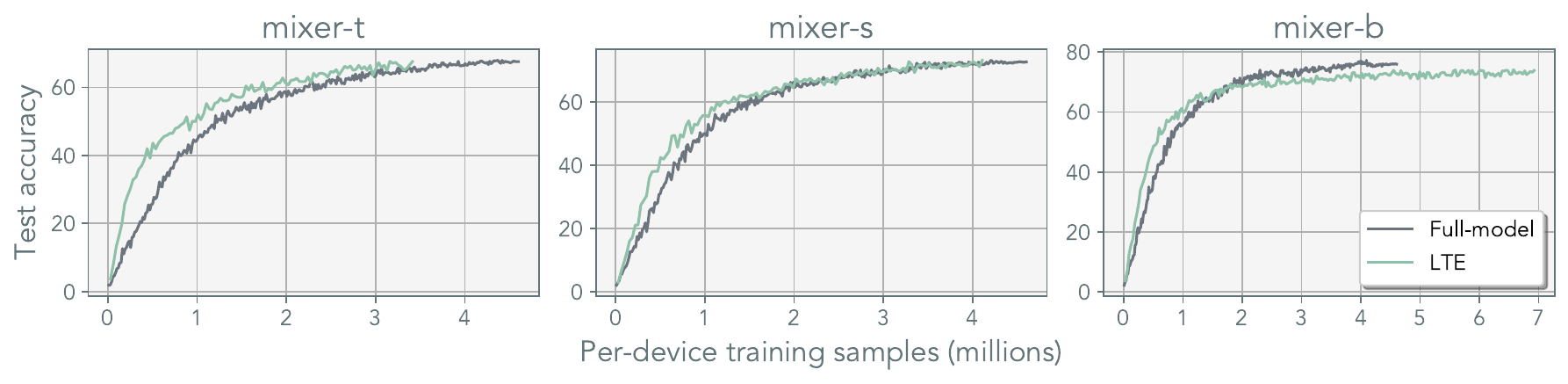}}
  \caption{\small \textbf{ImageNet100 classification on MLP-Mixer of varying scale}}
  \label{fig:both_figures}
\end{figure}

\newpage
\subsection{Language Modeling}
We also apply our method to language modeling. For these experiments, we utilized the nanoGPT codebase~\citep{karpathy2023nanogpt}. Detailed information about the architectures and datasets employed can be found in~\app{app:train-details}. Shakespeare's dataset was trained using MiniGPT~\citep{karpathy2023nanogpt}, while TinyStories~\citep{eldan2023tinystories} was trained on GPT2~\citep{radford2019language}. For Shakespeare, we used a configuration with rank \( r=16 \) and \( N=32 \) heads, and for TinyStories, we employed rank \( r=64 \) and \( N=32 \) heads. Consistent results were observed across all sizes. We observed on simple and small datasets that LTE has a beneficial regularization effect, reducing overfitting. \textit{Note}: We did not increase the training duration for these experiments and used fixed cumulative training samples.

\begin{figure}[htbp]
  \centering
\subfigure[LLM test error vs total training samples]{\label{fig:first}\includegraphics[width=0.49\linewidth]{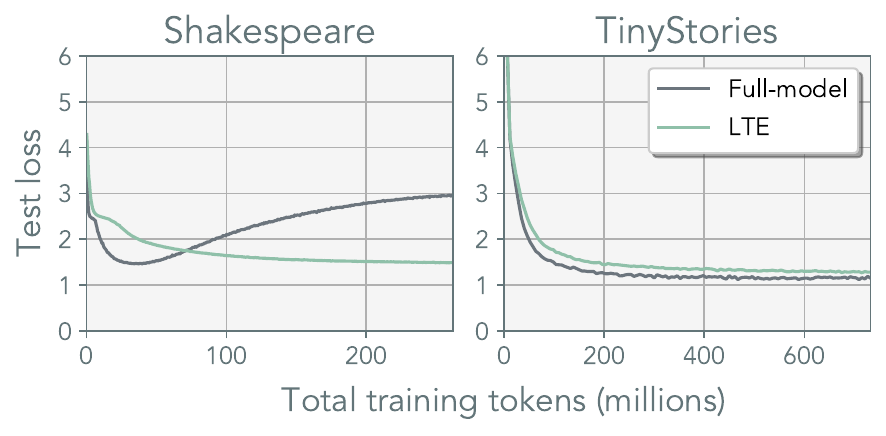}}
\subfigure[LLM test error vs training samples per device]{\label{fig:first}\includegraphics[width=0.49\linewidth]{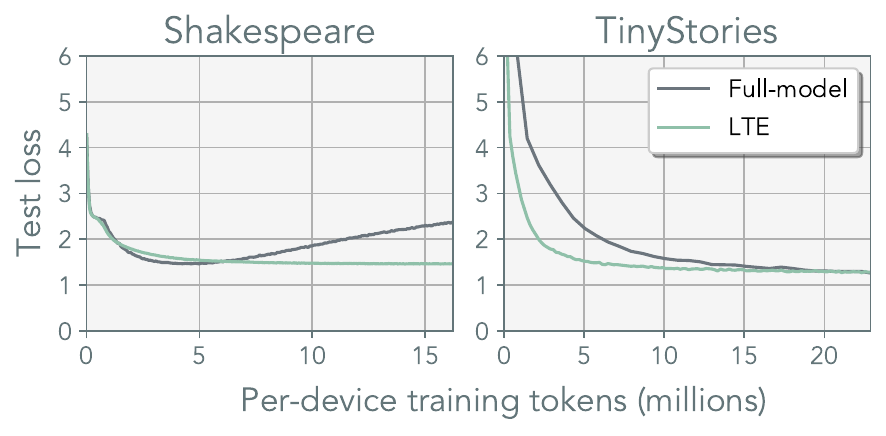}}
  \caption{\small \textbf{Additional LLM results on Shakespeare and TinyStories using GPT2}}
  \label{fig:both_figures}
\end{figure}

\end{document}